%% file: main.tex
\pdfoutput=1

\documentclass[11pt]{article}

\usepackage{colortbl}
\usepackage[table]{xcolor}
\usepackage[final]{acl} 
\usepackage{times}
\usepackage{latexsym}
\usepackage[T1]{fontenc}
\usepackage[utf8]{inputenc}
\usepackage{CJKutf8}

\usepackage{microtype}
\usepackage{inconsolata}

\usepackage{graphicx}
\usepackage{amsmath}
\usepackage{amssymb}
\usepackage{bbm}

\usepackage{pifont}
\usepackage{tabularx}
\usepackage{xspace}
\usepackage{makecell}
\usepackage{booktabs}
\usepackage{enumitem}
\usepackage{multirow}
\usepackage{adjustbox}
\usepackage[skins,breakable]{tcolorbox}
\usepackage{placeins}

\newcommand{\cmark}{\textcolor{green!60!black}{\ding{51}}} 
\newcommand{\xmark}{\textcolor{red}{\ding{55}}}

\definecolor{azure}{rgb}{0.94, 1.0, 1.0}
\definecolor{lightgray}{rgb}{0.9, 0.9, 0.9}
\definecolor{lightred}{rgb}{1.0, 0.94, 0.94}
\definecolor{decovecblueback}{RGB}{240, 248, 255} 
\definecolor{decovecblueframe}{RGB}{70, 130, 180}  
\definecolor{commentboxbg}{RGB}{255, 255, 255}

\title{CALM: Consistency-Aware Logit Masking in Decoding Space for Mitigating Overthinking of Large Reasoning Models}

\author{
  Shengjing Liu\thanks{~~Equal contribution.}, 
  Feiyang Li\footnotemark[1], 
  Yuxuan Yang, 
  Yile Wang\thanks{~~Corresponding author: \texttt{wangyile@szu.edu.cn}.}, 
  Hui Huang\\
  College of Computer Science and Software Engineering, Shenzhen University 
}

\begin{document}
\maketitle

\begin{abstract}
Large reasoning models (LRMs) have achieved strong performance on challenging mathematical and scientific tasks through long chain-of-thought reasoning, but they often suffer from overthinking with unnecessary reasoning steps. Existing truncation-based methods typically detect consecutive trial-answer agreement and force early exit, yet such local stopping signals do not necessarily indicate final-answer stability and can disrupt useful subsequent reflection or verification, potentially leading to performance degradation. We propose consistency-aware logit masking (\textsc{CALM}), a training-free decoding intervention for mitigating overthinking in LRMs. \textsc{CALM} first monitors trial answers at step boundaries and  then applies logit masking to action transition point (ATP) tokens such as “\textit{Wait}”, “\textit{Alternatively}”, and “\textit{Hmm}”, suppressing redundant transitional reasoning while allowing the model to continue its thought process more naturally. Experiments across four advanced LRMs and five benchmarks show that \textsc{CALM} reduces token consumption by 18.6\%–25.0\% compared with full CoT reasoning while maintaining performance, and can even improve accuracy by 0.07\%–1.48\% by reducing redundant reasoning. LLM-as-a-Judge results also demonstrate that \textsc{CALM} provides more coherent and concise results. Code is available at \url{https://github.com/szu-tera/CALM.git}.
\end{abstract}
\input{sec/1_introduction}

\input{sec/2_related_work}
\input{sec/3_preliminaries}

\input{sec/4_method}
\input{sec/5_experiments}

\input{sec/6_analysis}
\input{sec/7_conclusion}
\input{sec/8_limitation}

\bibliography{main}

\clearpage
\appendix
\input{sec/9_appendix}

\end{document}

%% file: sec/1_introduction.tex
\section{Introduction}
\label{sec:introduction}

Large reasoning models (LRMs) such as OpenAI o1~\cite{openai2024learning} and DeepSeek-R1~\cite{deepseekai2025deepseekr1} have advanced rapidly and achieved strong performance on solving challenging mathematical and graduate-level science problems, largely owing to their long chain-of-thought (CoT;~\citealp{wei2022chain}) reasoning and test-time scaling~\cite{snell2024scaling,muennighoff-etal-2025-s1}. However, research reveals that LRMs could exhibit \textit{overthinking}~\cite{su2025between,chen2025overthinking,cuadron2025danger}, characterized by extensive redundant reasoning that not only fails to improve model performance but also results in considerable token waste and extra inference latency~\cite{gema2025inverse,zhang-etal-2026-llms-really}. 

\begin{figure}[t!]
\centering
\includegraphics[width=\columnwidth]{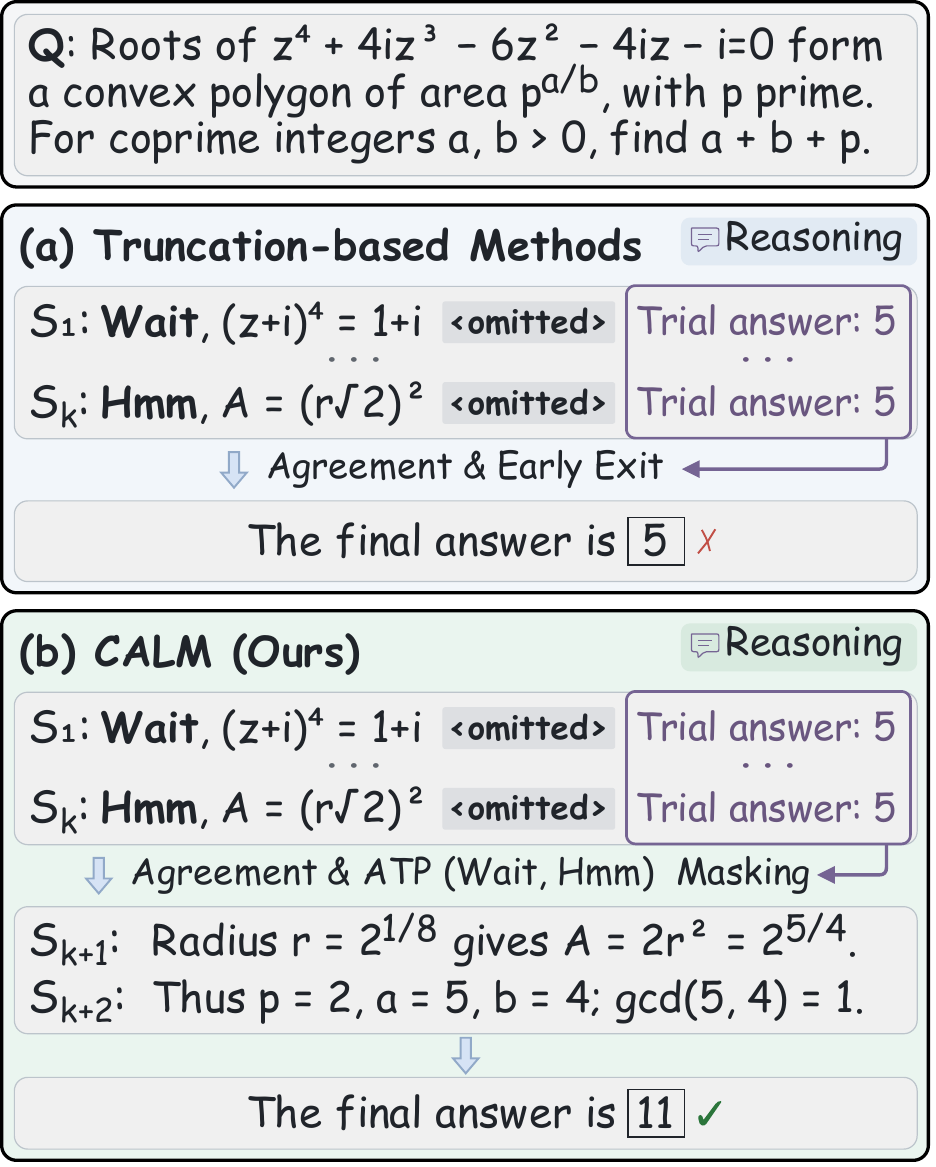}
\caption{Comparison of (a) truncation-based method and (b) our method that allows continual generation with action transition point masking after answer agreement.}
\label{fig:intro}
\end{figure}

\begin{table*}[t!]
\centering
\small
\setlength{\tabcolsep}{5pt}
\renewcommand{\arraystretch}{1.0}
\adjustbox{max width=\textwidth}{%
\begin{tabular}{@{}lccc@{}}
\toprule
\textbf{Related Work} & \textbf{Type} & \textbf{Signal} & \textbf{Action} \\
\midrule
L1~\cite{aggarwal2025l} & Reinforcement Learning & Length Penalty &  Reward Feedback\\
S-GRPO~\cite{dai2026sgrpo} & Reinforcement Learning & Step Sufficiency &  Reward Feedback\\
ThinkPrune~\cite{hou2026thinkprune} & Reinforcement Learning & Token Limit &  Reward Feedback\\
LightThinker~\cite{zhang-etal-2025-lightthinker} & Latent CoT Reasoning & Special Marker & Thought Compression \\
Step Entropy~\cite{li2026making} & Latent CoT Reasoning & Step Entropy & Thought Compression \\
\midrule
Chain of Draft~\cite{xu2025chainofdraft} & Prompt-based Constraint  & Draft Length & Context Awareness \\
Plan and Budget~\cite{lin2026plan} & Prompt-based Constraint  & Token Budget & Context Awareness \\
\midrule
Dynasor~\cite{fu2025dynasorcot} & Reasoning Truncation & Consecutive Answers & Terminate and Early Exit \\
Answer Conv.~\cite{liu-wang-2025-answer} & Reasoning Truncation & Consecutive Answers & Terminate and Early Exit \\
TRACE~\cite{li-etal-2026-efficient} & Reasoning Truncation & Temporal Consistency & Terminate and Early Exit \\
REFRAIN~\cite{sun2026stop} & Reasoning Truncation & Semantic Redundancy & Terminate and Early Exit \\
PUMA~\cite{min2026puma} & Reasoning Truncation & Semantic Redundancy & Answer with Token Limit \\
DEER~\cite{yang2025deer} & Reasoning Truncation & Transition Confidence & Answer with Token Limit \\
\textbf{\textsc{CALM} (this work)} & \textbf{Decoding Intervention} & \textbf{Consecutive Answers} & \textbf{Generate with Logit Masking} \\
\bottomrule
\end{tabular}%
}

\caption{Comparison of recent works on mitigating overthinking of large reasoning models.}
\label{tab:related}
\end{table*}

To reduce overthinking, recent efforts attempt to identify critical points during reasoning at which the model’s reasoning information is sufficient, so as to truncate the reasoning path by forcing the generation of the end-of-thinking token $\texttt{</think>}$ and eliciting the final answer directly. For example, researchers have used consecutive trial-answer agreement~\cite{fu2025dynasorcot,liu-wang-2025-answer} as a truncation signal, as shown in Figure~\ref{fig:intro}(a).

We argue that although this truncation-based early-exiting approach can reduce token consumption, it still restricts the model’s inherent reasoning ability. We conduct a pilot study ($\S$~\ref{sec:redundant_transitions}) and find that the stopping signal is not necessarily indicative of final-answer stability, with even considerable gaps observed on certain datasets. The reason can be that subsequent thinking may still contain steps that help improve the reasoning outcome, such as reflection and verification~\citep{deepseekai2025deepseekr1}. Meanwhile, interrupting the reasoning process severely suppresses the continuity of the model’s thinking, leading to incoherent reasoning context and a mismatch with the data and preferences encountered during training, thereby affecting the accuracy of the final answer~\citep{koh2026doesn, wang2026detection, wu2026are}. Therefore, we attempt to answer the following question: \textit{How can LRMs naturally stop thinking when their reasoning progress is detected to be approaching saturation?}

We begin with typical thinking words such as ``\textit{Wait}'', ``\textit{Alternatively}'', and ``\textit{Hmm}''~\cite{muennighoff-etal-2025-s1}, also termed as action transition point (ATP;~\citealp{yang2025deer}), and find that suppressing their generation probabilities can sustain generation while preserving accuracy ($\S$~\ref{sec:redundant_transitions}). The findings are also in line with recent work showing that ATP words do not genuinely reflect reasoning ability, but instead lead to lengthy reasoning steps~\cite{pandey2026selfdOUBT,liu2025understanding,yu2026revisiting}. Therefore, we propose consistency-aware logit masking (\textsc{CALM}), a simple yet effective inference-time decoding strategy for mitigating overthinking of LRMs, which consists of two  steps: detecting consecutive answer agreement ($\S$~\ref{subsec:consensus}) and action transition point logit masking ($\S$~\ref{subsec:masking}).

We evaluate on four advanced large reasoning models including DeepSeek-R1-Distill-Qwen (7B, 14B, 32B) and DeepSeek-R1-Distill-Llama-8B. Experiments across five mathematics and science benchmarks show that \textsc{CALM} outperforms recent baselines. In particular, CALM reduces token consumption by 18.6\%–25.0\% while maintaining performance compared with full CoT, and achieves a slight accuracy gain of 0.07\%–1.48\% through mitigating redundant reasoning.

In summary, our contributions are threefold:

1) Our pilot study shows that signals such as consecutive trial-answer agreement do not recover the completed-trace answer stably, revealing the limitations of truncation-based approaches, specifically their excessive reliance on local signals.

2) We propose \textsc{CALM}, a training-free decoding intervention method by suppressing the probability of typical thinking tokens to facilitate continual reasoning progress while reducing lengthy and redundant reasoning.

3) We evaluate two variants of \textsc{CALM} across four models and five benchmarks. Experimental results show that \textsc{CALM} substantially reduces redundant reasoning steps while outperforming recent baselines in terms of performance preservation.

%% file: sec/2_related_work.tex
\section{Related Work}
\label{sec:related}

\paragraph{Overthinking of LRMs.}
Recent work on overthinking of LRMs can be categorized into three types, as summarized in Table~\ref{tab:related}:

1) \textit{Training-based methods.}
Typical reinforcement learning methods include L1~\citep{aggarwal2025l}, which controls how long the model thinks through length penalties, S-GRPO~\citep{dai2026sgrpo}, which evaluates the sufficiency of intermediate reasoning steps for early exiting, and ThinkPrune~\citep{hou2026thinkprune}, which trains LRMs with an added token limit. For internal thought compression methods, LightThinker~\citep{zhang-etal-2025-lightthinker} compresses intermediate thoughts into compact representations, while \citet{li2026making} replaces selected low-entropy step with a special token $\texttt{[SKIP]}$. These methods incur prohibitive retraining overhead and cannot be deployed as plug-and-play interventions on frozen LRMs.

2) \textit{Prompt-based methods.}
Prompt-based methods encourage concise reasoning through prompt constraints. Chain of Draft~\citep{xu2025chainofdraft} limits reasoning to short draft steps. Plan and Budget~\citep{lin2026plan} assigns a length budget to the reasoning. Such constraints can be ignored on difficult problems or suppress necessary intermediate reasoning~\citep{yu2025premise}, so they are difficult to apply across diverse scenarios stably.

3) \textit{Inference-based methods.}
Reasoning truncation methods follow a ``detect, truncate, early exit'' pipeline. For example, Dynasor~\citep{fu2025dynasorcot} probes trial answers at fixed token intervals and stops when consecutive probes agree. Answer Convergence~\citep{liu-wang-2025-answer} probes at reasoning-chunk boundaries and exits when consecutive answers match.
TRACE~\citep{li-etal-2026-efficient} aggregates answer agreement and the confidence trajectory over a sliding window. REFRAIN~\citep{sun2026stop} identifies steps that are both reflective and redundant. PUMA~\citep{min2026puma} trains a detector to flag redundant steps and stops after verifying answer consistency and confidence. DEER~\citep{yang2025deer} elicits a trial answer at transitions and exits when confidence exceeds a threshold.  

These methods force the model to terminate its reasoning upon satisfaction of the stopping criterion, thereby restricting subsequent generation without token limitations. In contrast, we mitigate overthinking from the perspective of suppressing the probabilities of specific tokens in decoding space, yet maintain the capacity for sustained generation while reducing thinking redundancy.

\paragraph{Thinking Tokens in LRMs.}
\citet{muennighoff-etal-2025-s1} try appending additional thinking tokens such as “\textit{Wait}” and “\textit{Hmm}” to enable test-time scaling. DeepSeek-R1~\cite{deepseekai2025deepseekr1} also demonstrates an “aha moment” after reinforcement learning training, which appears as self-reflection following stop words such as “\textit{Wait}”. However,  \citet{liu2025understanding} show that the phenomenon already exists in base models and does not reflect reasoning ability. \citet{wang-etal-2025-wait} directly disables the generation of these words and find that it does not compromise the utility of multimodal models. \citet{yu2026revisiting} classifies such words as anthropomorphic reflection markers and finds that these markers primarily affect how reasoning trajectories are organized on the surface and sampled. These works mainly focus on empirical findings and mechanistic analysis, whereas we design a decoding strategy that conditionally suppresses such thinking tokens in long reasoning chains, retains the model’s capacity for continuous generation, and practically mitigates overthinking in LRMs.

%% file: sec/3_preliminaries.tex
\section{Preliminaries and Pilot Study}
\label{sec:prelim}

\subsection{Task Formulation and Notations}
\label{sec:task_formulation}
Given a question \(Q\), an autoregressive reasoning model generates a chain-of-thought reasoning sequence \(S=(s_1,\ldots,s_N)\), followed by a final answer \(y\). We segment the sequence into \(N\) steps at natural paragraph or step boundaries, with the rules given in Appendix~\ref{sec:step_seg}. The task is to reduce unnecessary redundant reasoning steps while preserving the accuracy of  answer \(y\). Let \(s_{1:n}\) denote the reasoning prefix up to step \(n\). Following the probing paradigm of prior early-exit studies~\cite{fu2025dynasorcot,liu-wang-2025-answer}, we sample an intermediate trial answer \(a_n\) conditioned on \(s_{1:n}\) at each step boundary \(n\) by appending an answer-inducing suffix on an independent decoding branch. This probing branch does not alter the original reasoning context. Consequently, each reasoning step \(s_n\) is associated with a trial answer \(a_n\). The suffix strings are provided in Appendix~\ref{sec:answer_suffix}.

Tracking intermediate trial answers enables monitoring reasoning stability during generation. Notably, existing early-exit methods adopt consecutive trial answer agreement as a criterion to early-stop and truncate generation~\cite{fu2025dynasorcot,liu-wang-2025-answer}.

\subsection{Action Transition Point}
During long chain-of-thought generation, LRMs alternate among distinct thinking actions, such as approach exploration and procedural execution. Switches between these actions are typically indicated by explicit thinking tokens such as ``\emph{Wait}''. Following the naming by \citet{yang2025deer}, we formalize these tokens as action transition points (ATP) and denote their set by \(\mathcal{V}_{\mathrm{ATP}}\). 

There is currently no rigorous definition of the ATP vocabulary. Following \citet{wang-etal-2025-wait}, we use three LRMs on five benchmarks to identify the sentence-initial words of different reasoning steps, and select the top 8 words to form the initial \(\mathcal{V}^{\rm init}_{\mathrm{ATP}}\) = \{``\textit{So}'',``\textit{Wait}'',  ``\textit{But}'',``\textit{Therefore}'',  ``\textit{Alternatively}'',  ``\textit{However}'', ``\textit{Maybe}'',``\textit{Hmm}''\}. We find that reflection words such as ``\textit{Wait}'', ``\textit{Alternatively}'', and ``\textit{Hmm}'' have been discussed in recent work~\cite{muennighoff-etal-2025-s1,wang2025demystifying,yu2026revisiting} and these words may be highly correlated with redundant reasoning, and other control words regulate verification and thinking, thereby preserving the accuracy of long chain-of-thought reasoning. We therefore focus on three words \(\mathcal{V}_{\mathrm{ATP}}\) = \{``\textit{Wait}'', ``\textit{Alternatively}'', and ``\textit{Hmm}''\}, and we conduct vocabulary ablation in Section~\ref{subsec:ablation}.

\subsection{Reasoning Trace after Answer Agreement}
\label{sec:redundant_transitions}
We hypothesize that once tentative answer agreement is reached, subsequent ATP triggers predominantly degenerate into computational redundancy, needlessly bloating the reasoning trajectory. To examine this, we track the reasoning continuation starting from the first answer agreement. We conduct a pilot study on DeepSeek-R1-Distill-Qwen-7B~\cite{deepseekai2025deepseekr1} and OlympiadBench~\cite{he2024olympiadbench} to investigate how accuracy changes with the mean number of continuation tokens after consecutive identical answers, comparing unconstrained vanilla CoT with a continuation that suppresses \(\mathcal{V}_{\mathrm{ATP}}\) tokens after agreement. Detailed settings are given in Appendix~\ref{sec:identical_continuation}.

\begin{figure}[t!]
\centering
\includegraphics[width=\columnwidth]{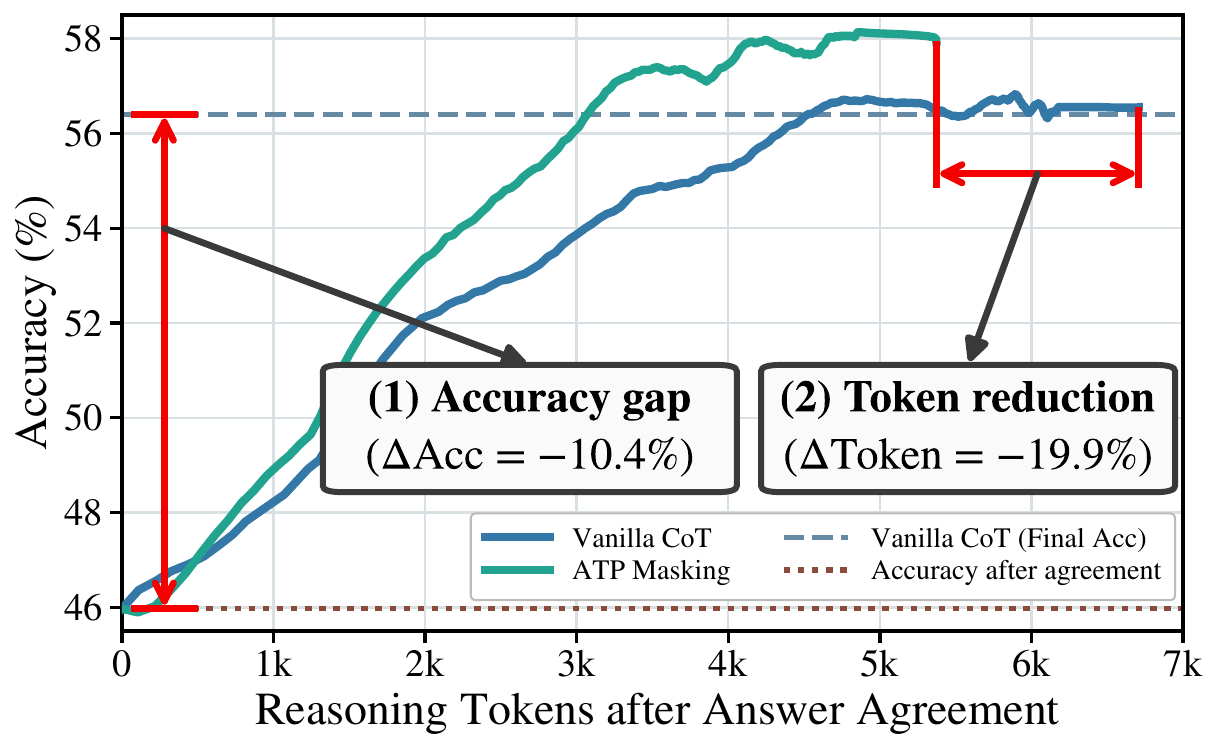}

\caption{Pilot study on accuracy and tokens on continued reasoning trace after trial answer agreement for DeepSeek-R1-Distill-Qwen-7B on OlympiadBench.}

\label{fig:post_agreement_dynamics}
\end{figure}

Figure~\ref{fig:post_agreement_dynamics} illustrates two key dynamics of this continuation. First, immediate truncation upon identical trial answers incurs a substantial accuracy gap. At the start of the continuation (\(x=0\)), the probed trial answer achieves an accuracy of only 46.0\%, \textbf{falling 10.4 points below vanilla CoT} (56.4\%). Truncating immediately at this boundary severs downstream derivations and discards the opportunity for later revisions. Second, both continuations allow the model to conclude, while suppressing ATP tokens shortens the continuation. Suppressing \(\mathcal{V}_{\mathrm{ATP}}\) after agreement \textbf{saves 19.9\% of post-agreement tokens} while reaching a final accuracy comparable to Vanilla CoT.

%% file: sec/4_method.tex
\section{Method}
\label{sec:method}
\begin{figure*}[t!]
\centering
\includegraphics[width=\textwidth]{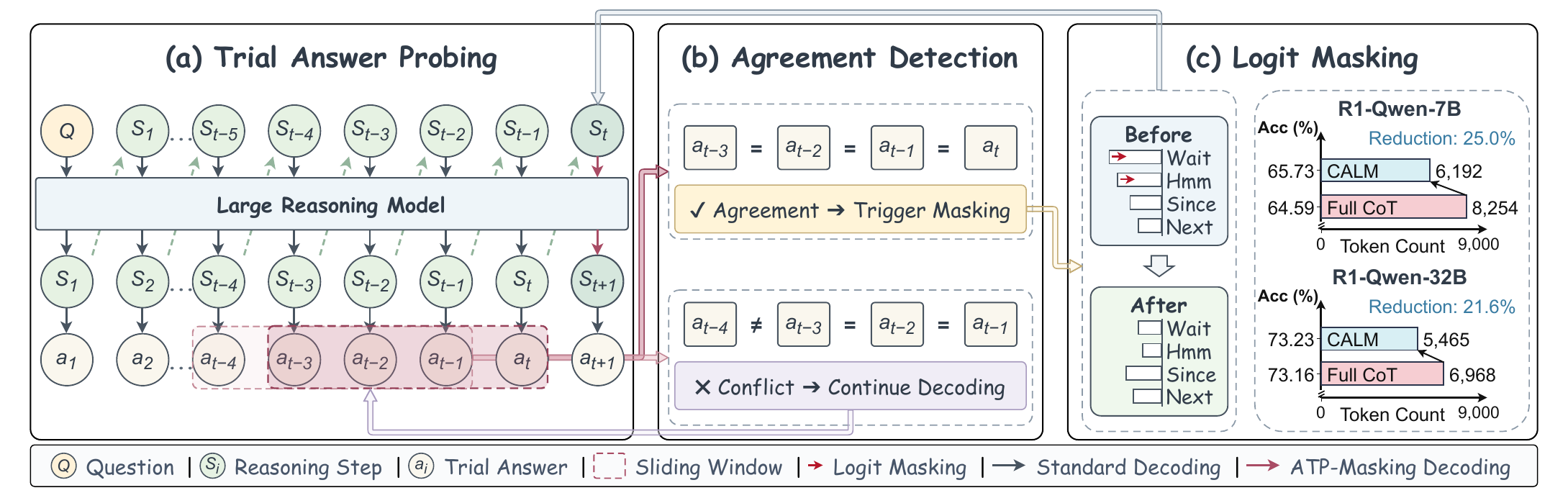}
\caption{The overall pipeline of \textsc{CALM}. A trial answer is probed at each step boundary. After the same trial answer remains identical for \(k\) consecutive steps, \textsc{CALM} lowers the logits of ATP tokens and continues reasoning.}
\label{fig:framework}
\end{figure*}

Grounded in the empirical findings in Section~\ref{sec:redundant_transitions}, we propose consistency-aware logit masking (\textsc{CALM}), a training-free decoding-based method for mitigating overthinking.Rather than terminating generation early by truncating, \textsc{CALM} allows generation after detecting consecutive trial-answer agreement. Figure~\ref{fig:framework} illustrates the overall pipeline.

\subsection{Detecting Consecutive Answer Agreement}
\label{subsec:consensus}

At each step boundary \(n\), \textsc{CALM} obtains the current trial answer \(a_n\) following the probing procedure described in Section~\ref{sec:prelim}.
We monitor the sequence of trial answers to detect the earliest window in which the model outputs identical answers for \(k\) consecutive steps.
Formally, the consecutive-agreement indicator \(c_n\) and the agreement boundary step \(b^{\star}\) are defined as:
\begin{equation}
c_{n} = \mathbbm{1}[a_{n}=a_{n-1}=\cdots=a_{n-k+1}],
\label{eq:consensus}
\end{equation}
\begin{equation}
b^{\star} = \min\{n\geq {\rm MSS} : c_n=1\},
\label{eq:boundary}
\end{equation}
where $\mathbbm{1}$[$\cdot$] denotes the indicator function and $\rm {MSS}$ is the minimum stop step. If the trial answers never reach \(k\) consecutive matches during generation, then \(b^{\star}=\infty\) and the reasoning trajectory remains entirely unmodified.

\subsection{Action Transition Point Logit Masking}
\label{subsec:masking}
After obtaining the agreement boundary \(b^{\star}\), \textsc{CALM} adds a masking bias to the logits of candidate tokens that match \(\mathcal{V}_{\mathrm{ATP}}\). Formally, at decoding position \(t\) within the reasoning steps, let \(h_t\) denote the final hidden state and \(W_{\mathrm{LM}}\) the vocabulary projection weight of the LM Head. Let $\mathcal{V}$ denote the model's token vocabulary. For each candidate token \(v\in\mathcal{V}\), the raw logit is \(\ell_t(v)=[W_{\mathrm{LM}}h_t]_{{\rm idx}:v}\), to which we add a bias \(b_t(v)\) for logit masking:
\begin{equation}
\begin{aligned}
\ell_t(v) &= [W_{\mathrm{LM}}h_t]_{{\rm idx}:v}, \\
\tilde{\ell}_{t}(v) &= \ell_t(v) + b_t(v),
\end{aligned}
\label{eq:mask}
\end{equation}
where ${{\rm idx}:v}$ denotes the index of \(v \in\mathcal{V}\). Let \(t\) be a token position and \(n(t)\) be the corresponding reasoning step. We investigate two masking strategies for the bias term \(b_t(v)\):

\textbf{Hard Masking} employs a constant value to strictly prohibit ATP tokens once the agreement boundary is crossed:
\begin{equation}
b_t(v) =
\begin{cases}
-\infty & \text{if } n(t)>b^{\star},\, v\in\mathcal{B}_t, \\
0 & \text{otherwise},
\end{cases}
\label{eq:bias}
\end{equation}
where \(\mathcal{B}_t\subseteq\mathcal{V}\) is the set of token ids that complete an ATP surface form under the current prefix. The surface forms and the matching rules are given in Appendix~\ref{app:atp_matching}. Vocabulary ablations are reported in Section~\ref{subsec:ablation}.

\textbf{Soft Masking} adopts an adaptive progressive penalty through decisive damping rather than a rigid ban. Let \(c(t)\) denote the occurrence count of ATP strings completed after \(b^{\star}\) and before position \(t\), with penalty \(\eta(c)=-10(1-\rho^{c})\) for \(0<\rho<1\):
\begin{equation}
b_t(v) =
\begin{cases}
\eta(c(t)) & \text{if } n(t)>b^{\star},\, v\in\mathcal{B}_t, \\
0 & \text{otherwise},
\end{cases}
\label{eq:bias_count}
\end{equation}
where the penalty is \(0\) when \(c(t)=0\), and smoothly approaches \(-10\) as \(c(t)\) accumulates. Under a softmax operation, a value of \(-10\) reduces an affected token's relative weight by a factor of about \(e^{-10}\approx 4.5\times 10^{-5}\). The sensitivity of hyperparameter $\rho$ is shown in Section~\ref{subsec:rho}.

After applying probability suppression to ATP tokens, \textsc{CALM} enables continual decoding without truncation or imposing additional token limits. We evaluate both \textsc{CALM} (soft) with progressive penalty and \textsc{CALM} (hard) with constant masking in our main experiments.

%% file: sec/5_experiments.tex
\section{Experiments}
\label{sec:experiments}

\begin{table*}[t!]
\centering
\definecolor{tokrose}{RGB}{191,0,64}
\newcommand{\tokc}[2]{\cellcolor{tokrose!#1}#2}
\DeclareFontShape{OT1}{cmr}{m}{n}{%
  <-6>cmr5<6-7>cmr6<7-8>cmr7<8-9>cmr8<9-10>cmr9%
  <10-12>cmr10<12-17>cmr12<17->cmr17}{}
\DeclareFontShape{OML}{cmm}{m}{it}{%
  <-6>cmmi5<6-7>cmmi6<7-8>cmmi7<8-9>cmmi8<9-10>cmmi9%
  <10-12>cmmi10<12->cmmi12}{}
\DeclareFontShape{OMS}{cmsy}{m}{n}{%
  <-6>cmsy5<6-7>cmsy6<7-8>cmsy7<8-9>cmsy8<9-10>cmsy9<10->cmsy10}{}
\DeclareFontShape{U}{lasy}{m}{n}{%
  <-6>lasy5<6-7>lasy6<7-8>lasy7<8-9>lasy8<9-10>lasy9<10->lasy10}{}
\fontsize{6.5pt}{7.8pt}\selectfont
\setlength{\tabcolsep}{0.7pt}
\renewcommand{\arraystretch}{1.05}
\begin{tabular*}{\textwidth}{@{\extracolsep{\fill}}l*{13}{c}@{}}
\toprule
\multirow{2}{*}{\textbf{Method}}
 & \multicolumn{2}{c}{\textbf{MATH-500}}
 & \multicolumn{2}{c}{\textbf{Olympiad}}
 & \multicolumn{2}{c}{\textbf{GPQA}}
 & \multicolumn{2}{c}{\textbf{AIME25}}
 & \multicolumn{2}{c}{\textbf{AMC23}}
 & \multicolumn{3}{c}{\textbf{Overall}} \\
\cmidrule(lr){2-3}\cmidrule(lr){4-5}\cmidrule(lr){6-7}\cmidrule(lr){8-9}\cmidrule(lr){10-11}\cmidrule(lr){12-14}
 & Acc$\uparrow$ & Tok$\downarrow$
 & Acc$\uparrow$ & Tok$\downarrow$
 & Acc$\uparrow$ & Tok$\downarrow$
 & Acc$\uparrow$ & Tok$\downarrow$
 & Acc$\uparrow$ & Tok$\downarrow$
 & Acc$\uparrow$ & Tok$\downarrow$ & TE$\uparrow$ \\
\midrule
\multicolumn{14}{c}{\textit{\textbf{DeepSeek-R1-Distill-Qwen-7B}}} \\
Full CoT & 92.33 & 3,847 & 57.63 & 8,718 & 47.14 & 7,327 & 36.67 & 15,080 & 89.17 & 6,297 & 64.59 & 8,254 & 58.57 \\
Ans.\ Conv. & 63.60 & \tokc{74}{1,006 (-73.8\%)} & 31.46 & \tokc{80}{1,490 (-82.9\%)} & 26.77 & \tokc{80}{348 (-95.3\%)} & 14.44 & \tokc{80}{3,020 (-80.0\%)} & 44.17 & \tokc{79}{1,293 (-79.5\%)} & 36.09 & \tokc{80}{1,432 (-82.7\%)} & 34.75 \\
Dynasor & 85.13 & \tokc{31}{2,655 (-31.0\%)} & 51.41 & \tokc{33}{5,820 (-33.2\%)} & 34.18 & \tokc{75}{1,802 (-75.4\%)} & 34.44 & \tokc{12}{13,202 (-12.5\%)} & 81.67 & \tokc{31}{4,348 (-31.0\%)} & 57.37 & \tokc{33}{5,565 (-32.6\%)} & 52.95 \\
TRACE & 89.33 & \tokc{28}{2,764 (-28.2\%)} & 53.53 & \tokc{50}{4,340 (-50.2\%)} & 43.77 & \tokc{67}{2,410 (-67.1\%)} & 34.44 & \tokc{45}{8,270 (-45.2\%)} & 77.50 & \tokc{38}{3,936 (-37.5\%)} & 59.71 & \tokc{47}{4,344 (-47.4\%)} & 56.10 \\
PUMA & 90.93 & \tokc{32}{2,599 (-32.4\%)} & 56.10 & \tokc{41}{5,115 (-41.3\%)} & 47.31 & \tokc{8}{6,768 (-7.6\%)} & 36.67 & \tokc{18}{12,366 (-18.0\%)} & 84.17 & \tokc{27}{4,594 (-27.0\%)} & 63.03 & \tokc{24}{6,288 (-23.8\%)} & 58.15 \\
DEER & 89.80 & \tokc{37}{2,436 (-36.7\%)} & 56.94 & \tokc{25}{6,545 (-24.9\%)} & \underline{49.16} & 7,406 (+1.1\%) & 36.67 & \tokc{19}{12,210 (-19.0\%)} & \underline{90.00} & \tokc{26}{4,666 (-25.9\%)} & 64.51 & \tokc{19}{6,653 (-19.4\%)} & 59.41 \\
\textsc{CALM} (Soft) & \textbf{92.47} & \tokc{18}{3,152 (-18.1\%)} & \textbf{58.52} & \tokc{18}{7,162 (-17.8\%)} & \textbf{50.51} & \tokc{26}{5,393 (-26.4\%)} & \underline{40.00} & \tokc{19}{12,288 (-18.5\%)} & \textbf{90.83} & \tokc{27}{4,593 (-27.1\%)} & \textbf{66.47} & \tokc{21}{6,518 (-21.0\%)} & \textbf{61.45} \\
\textsc{CALM} (Hard) & \underline{92.27} & \tokc{26}{2,843 (-26.1\%)} & \underline{57.33} & \tokc{24}{6,663 (-23.6\%)} & 48.48 & \tokc{31}{5,059 (-31.0\%)} & \textbf{42.22} & \tokc{22}{11,749 (-22.1\%)} & 88.33 & \tokc{26}{4,646 (-26.2\%)} & \underline{65.73} & \tokc{25}{6,192 (-25.0\%)} & \underline{60.91} \\
\midrule
\multicolumn{14}{c}{\textit{\textbf{DeepSeek-R1-Distill-Qwen-14B}}} \\
Full CoT & 92.87 & 3,504 & 62.67 & 7,792 & 50.00 & 6,335 & 53.33 & 12,296 & 94.17 & 5,448 & 70.61 & 7,075 & 65.26 \\
Ans.\ Conv. & 58.13 & \tokc{76}{839 (-76.1\%)} & 26.86 & \tokc{80}{1,125 (-85.6\%)} & 38.22 & \tokc{80}{330 (-94.8\%)} & 12.22 & \tokc{79}{2,606 (-78.8\%)} & 40.00 & \tokc{79}{1,165 (-78.6\%)} & 35.09 & \tokc{80}{1,213 (-82.9\%)} & 33.95 \\
Dynasor & 91.40 & 3,573 (+2.0\%) & 61.48 & \tokc{8}{7,162 (-8.1\%)} & 49.33 & \tokc{44}{3,577 (-43.5\%)} & 47.78 & 12,765 (+3.8\%) & \underline{93.33} & 5,606 (+2.9\%) & 68.66 & \tokc{8}{6,537 (-7.6\%)} & 63.63 \\
TRACE & 89.27 & \tokc{27}{2,555 (-27.1\%)} & 56.79 & \tokc{48}{4,021 (-48.4\%)} & 42.26 & \tokc{66}{2,135 (-66.3\%)} & 40.00 & \tokc{39}{7,549 (-38.6\%)} & 84.17 & \tokc{28}{3,914 (-28.2\%)} & 62.50 & \tokc{43}{4,035 (-43.0\%)} & 59.11 \\
PUMA & \underline{92.07} & \tokc{23}{2,692 (-23.2\%)} & 60.64 & \tokc{32}{5,283 (-32.2\%)} & 53.03 & \tokc{14}{5,419 (-14.5\%)} & 51.11 & \tokc{9}{11,141 (-9.4\%)} & \textbf{94.17} & \tokc{16}{4,566 (-16.2\%)} & 70.20 & \tokc{18}{5,820 (-17.7\%)} & 65.61 \\
DEER & 90.87 & \tokc{30}{2,465 (-29.7\%)} & 60.94 & \tokc{19}{6,349 (-18.5\%)} & 54.38 & 6,381 (+0.7\%) & 51.11 & \tokc{2}{12,000 (-2.4\%)} & 92.50 & \tokc{18}{4,470 (-18.0\%)} & 69.96 & \tokc{10}{6,333 (-10.5\%)} & 65.05 \\
\textsc{CALM} (Soft) & \textbf{93.33} & \tokc{18}{2,883 (-17.7\%)} & \underline{61.88} & \tokc{19}{6,275 (-19.5\%)} & \textbf{57.41} & \tokc{22}{4,962 (-21.7\%)} & \textbf{57.78} & \tokc{7}{11,421 (-7.1\%)} & \underline{93.33} & \tokc{12}{4,768 (-12.5\%)} & \textbf{72.75} & \tokc{14}{6,062 (-14.3\%)} & \textbf{68.01} \\
\textsc{CALM} (Hard) & 92.00 & \tokc{21}{2,769 (-21.0\%)} & \textbf{61.98} & \tokc{22}{6,083 (-21.9\%)} & \underline{56.73} & \tokc{27}{4,635 (-26.8\%)} & \underline{55.56} & \tokc{9}{11,184 (-9.0\%)} & \textbf{94.17} & \tokc{24}{4,121 (-24.4\%)} & \underline{72.09} & \tokc{19}{5,758 (-18.6\%)} & \underline{67.54} \\
\midrule
\multicolumn{14}{c}{\textit{\textbf{DeepSeek-R1-Distill-Qwen-32B}}} \\
Full CoT & 91.40 & 2,834 & 62.57 & 7,221 & 63.80 & 6,101 & 52.22 & 12,843 & 95.83 & 5,841 & 73.16 & 6,968 & 67.87 \\
Ans.\ Conv. & 64.60 & \tokc{79}{598 (-78.9\%)} & 27.85 & \tokc{80}{834 (-88.5\%)} & 44.28 & \tokc{80}{277 (-95.5\%)} & 7.78 & \tokc{80}{1,433 (-88.8\%)} & 41.67 & \tokc{80}{923 (-84.2\%)} & 37.23 & \tokc{80}{813 (-88.3\%)} & 36.45 \\
Dynasor & 88.67 & 2,833 (0.0\%) & 60.44 & \tokc{13}{6,280 (-13.0\%)} & 54.88 & \tokc{52}{2,947 (-51.7\%)} & 50.00 & 12,899 (+0.4\%) & 93.33 & \tokc{2}{5,739 (-1.7\%)} & 69.47 & \tokc{12}{6,140 (-11.9\%)} & 64.68 \\
TRACE & 88.87 & \tokc{26}{2,110 (-25.5\%)} & 56.54 & \tokc{50}{3,641 (-49.6\%)} & 55.39 & \tokc{65}{2,128 (-65.1\%)} & 43.33 & \tokc{49}{6,563 (-48.9\%)} & 86.67 & \tokc{41}{3,427 (-41.3\%)} & 66.16 & \tokc{49}{3,574 (-48.7\%)} & 63.10 \\
PUMA & 90.80 & \tokc{18}{2,338 (-17.5\%)} & 60.94 & \tokc{22}{5,645 (-21.8\%)} & 63.97 & \tokc{10}{5,475 (-10.3\%)} & \textbf{53.33} & \tokc{14}{11,011 (-14.3\%)} & 92.50 & \tokc{14}{5,022 (-14.0\%)} & 72.31 & \tokc{15}{5,898 (-15.4\%)} & 67.67 \\
DEER & \underline{90.87} & \tokc{21}{2,232 (-21.2\%)} & \underline{62.07} & \tokc{19}{5,850 (-19.0\%)} & \textbf{65.15} & 6,244 (+2.3\%) & 48.89 & \tokc{11}{11,396 (-11.3\%)} & \underline{95.00} & \tokc{19}{4,739 (-18.9\%)} & 72.40 & \tokc{13}{6,092 (-12.6\%)} & 67.64 \\
\textsc{CALM} (Soft) & \textbf{91.33} & \tokc{16}{2,387 (-15.8\%)} & \textbf{62.52} & \tokc{19}{5,885 (-18.5\%)} & 64.65 & \tokc{22}{4,729 (-22.5\%)} & 51.11 & \tokc{14}{11,109 (-13.5\%)} & \underline{95.00} & \tokc{21}{4,613 (-21.0\%)} & \underline{72.92} & \tokc{18}{5,745 (-17.6\%)} & \underline{68.38} \\
\textsc{CALM} (Hard) & \textbf{91.33} & \tokc{20}{2,258 (-20.3\%)} & 61.98 & \tokc{21}{5,695 (-21.1\%)} & \underline{64.81} & \tokc{24}{4,620 (-24.3\%)} & \underline{52.22} & \tokc{18}{10,581 (-17.6\%)} & \textbf{95.83} & \tokc{29}{4,173 (-28.6\%)} & \textbf{73.23} & \tokc{22}{5,465 (-21.6\%)} & \textbf{68.87} \\
\midrule
\multicolumn{14}{c}{\textit{\textbf{DeepSeek-R1-Distill-Llama-8B}}} \\
Full CoT & 73.40 & 3,558 & 48.20 & 8,161 & 16.50 & 7,372 & 30.00 & 13,999 & 81.67 & 6,487 & 49.95 & 7,915 & 44.12 \\
Ans.\ Conv. & 59.60 & \tokc{74}{910 (-74.4\%)} & 28.49 & \tokc{80}{1,445 (-82.3\%)} & \textbf{27.78} & \tokc{80}{543 (-92.6\%)} & 11.11 & \tokc{80}{1,737 (-87.6\%)} & 35.83 & \tokc{80}{1,063 (-83.6\%)} & 32.56 & \tokc{80}{1,140 (-85.6\%)} & 31.48 \\
Dynasor & 67.07 & \tokc{22}{2,788 (-21.6\%)} & 41.93 & \tokc{39}{4,990 (-38.9\%)} & 22.56 & \tokc{70}{2,195 (-70.2\%)} & 22.22 & \tokc{18}{11,489 (-17.9\%)} & 68.33 & \tokc{28}{4,657 (-28.2\%)} & 44.42 & \tokc{34}{5,224 (-34.0\%)} & 40.22 \\
TRACE & 69.87 & \tokc{28}{2,546 (-28.4\%)} & 46.57 & \tokc{51}{4,012 (-50.8\%)} & \underline{27.10} & \tokc{61}{2,860 (-61.2\%)} & 22.22 & \tokc{58}{5,824 (-58.4\%)} & 70.00 & \tokc{36}{4,177 (-35.6\%)} & 47.15 & \tokc{51}{3,884 (-50.9\%)} & 43.87 \\
PUMA & 71.53 & \tokc{23}{2,751 (-22.7\%)} & 45.68 & \tokc{33}{5,436 (-33.4\%)} & 25.42 & \tokc{6}{6,942 (-5.8\%)} & 27.78 & \tokc{9}{12,697 (-9.3\%)} & \textbf{77.50} & \tokc{22}{5,063 (-22.0\%)} & 49.58 & \tokc{17}{6,578 (-16.9\%)} & 44.53 \\
DEER & 72.13 & \tokc{11}{3,163 (-11.1\%)} & \textbf{47.51} & \tokc{4}{7,821 (-4.2\%)} & 16.84 & \tokc{2}{7,246 (-1.7\%)} & \textbf{32.22} & \tokc{9}{12,728 (-9.1\%)} & \underline{75.83} & 6,582 (+1.5\%) & 48.91 & \tokc{5}{7,508 (-5.1\%)} & 43.31 \\
\textsc{CALM} (Soft) & \textbf{75.00} & \tokc{18}{2,930 (-17.7\%)} & \underline{47.36} & \tokc{19}{6,586 (-19.3\%)} & 24.92 & \tokc{23}{5,659 (-23.2\%)} & \underline{28.89} & \tokc{14}{11,986 (-14.4\%)} & 73.33 & \tokc{21}{5,139 (-20.8\%)} & \underline{49.90} & \tokc{18}{6,460 (-18.4\%)} & \underline{44.92} \\
\textsc{CALM} (Hard) & \underline{74.07} & \tokc{20}{2,844 (-20.1\%)} & 46.96 & \tokc{20}{6,489 (-20.5\%)} & 25.25 & \tokc{28}{5,313 (-27.9\%)} & \underline{28.89} & \tokc{18}{11,412 (-18.5\%)} & \underline{75.83} & \tokc{21}{5,157 (-20.5\%)} & \textbf{50.20} & \tokc{21}{6,243 (-21.1\%)} & \textbf{45.35} \\
\bottomrule
\end{tabular*}
\caption{Averaged results using three seeds across DeepSeek-R1 distillation models. Acc: accuracy. Tok: average token count. TE: token efficiency. Values in parentheses are token changes relative to Full CoT. The best and second-best accuracy and TE values among the efficient-reasoning methods are shown in \textbf{bold} and \underline{underlined}, respectively; Full CoT is excluded from this ranking.}
\label{tab:main}
\end{table*}

\subsection{Experimental Setup}
\label{subsec:setup}

\paragraph{Benchmarks.}
We evaluate on five reasoning benchmarks that span several difficulty levels: four mathematical benchmarks including MATH-500~\cite{lightman_let_2024_math500}, OlympiadBench~\cite{he2024olympiadbench}, AIME 2025~\cite{mathai2025aime}, and AMC 2023~\cite{AMC23}, and one scientific reasoning benchmark:GPQA-Diamond~\cite{rein2024gpqa}.
\paragraph{Models.}
Four DeepSeek models are used in main experiments, including DeepSeek-R1-Distill-Qwen (7B, 14B, and 32B)~\cite{deepseekai2025deepseekr1} and DeepSeek-R1-Distill-Llama-8B.
For additional models Llama-3.1-Nemotron-Nano-8B~\cite{bercovich2025llamnemotron}, Qwen3 (4B, 8B, 30B-A3B, 32B)~\cite{qwen3technicalreport}, and QwQ-32B~\cite{qwq32b}, results are given in Appendix~\ref{app:more_models}.

\paragraph{Baselines.}
To make a fair evaluation, we compare with inference-based methods including \textbf{Answer Convergence} (Ans. Conv.;~\citealp{liu-wang-2025-answer}), \textbf{Dynasor}~\cite{fu2025dynasorcot}, \textbf{TRACE}~\cite{li-etal-2026-efficient}, \textbf{PUMA}~\cite{min2026puma}, and \textbf{DEER}~\cite{yang2025deer}. We obtain and report the results based on their open-source version.

\paragraph{Evaluation Metrics.}
We report accuracy (Acc) and average generated tokens (Tok).
Overall, we additionally report token efficiency (TE), using the OckScore~\cite{du2025ockbench} defined as accuracy minus \(\lambda\ln(T/C+1)\), with \(\lambda=10\), \(C=10{,}000\), natural logarithm, accuracy in percentage points, and \(T\) the average token count.

\paragraph{Implementation Details.}
All experiments use 4 NVIDIA L40 GPUs, and all methods run under one vLLM host. The main generation budget is \(32{,}768\) tokens. If this budget ends before the thought closes, the answer continuation is at most \(2{,}048\) tokens. Each reported cell is the mean over seeds 42, 0, and 1.
The main experiments set the window length to \(k=4\), and the minimum stop step is \(\mathrm{MSS}=10\), so \(b^{\star}\) is no earlier than step 10.
Section~\ref{subsec:window} compares other lengths. We provide a more detailed analysis at Appendix~\ref{sec:instruction_decoding}.

\subsection{Main Results}
\label{sec:main_results}
\suppressfloats[t]

\paragraph{Overall Performance.}
As shown in Table~\ref{tab:main}, \textsc{CALM} achieves the highest overall TE on all four backbones. Relative to Full CoT, \textsc{CALM} (Hard) reduces overall token usage by \(18.6\%\)--\(25.0\%\), with overall accuracy \(0.07\)--\(1.48\) points higher. It also achieves higher overall accuracy with fewer tokens than both PUMA and DEER on every backbone. Both \textsc{CALM} variants reduce token usage across all 20 model--benchmark settings, although their effects on accuracy vary across tasks. Table~\ref{tab:more_models} in Appendix~\ref{app:more_models} also presents results on six additional models, where \textsc{CALM} similarly reduces token consumption by \(14\%\)--\(18.9\%\) and achieves better accuracy and token efficiency than other baselines.

\paragraph{Comparison with Early-Exit Methods.}
Early-exit baselines exhibit different accuracy--efficiency trade-offs. Answer Convergence reduces overall token usage by \(82.7\%\)--\(88.3\%\), but lowers overall accuracy by \(17.39\)--\(35.93\) percentage points relative to Full CoT. In contrast, DEER increases token usage by \(0.7\%\)--\(2.3\%\) on GPQA-Diamond for the three Qwen backbones, because its high threshold almost never triggers early exit on GPQA-Diamond and trial-answer probing adds tokens. In Appendix~\ref{app:stop_coverage}, we provide detailed statistics on the proportion of this early-stopping signal (stop coverage) and find that \textsc{CALM} intervenes on more questions than PUMA and DEER truncate in most settings, which helps explain the larger token reductions on some benchmarks. Overall, the superiority of \textsc{CALM} comes from the suppression of ATP after consecutive identical answers, while subsequent deduction and correction continue without degrading intermediate reasoning quality.

\paragraph{Soft vs.\ Hard Masking.}
Hard masking achieves lower overall token usage than soft masking on all four backbones, whereas their accuracy advantage depends on the backbone. Soft masking attains the highest overall accuracy on Qwen-7B and Qwen-14B, reaching \(66.47\%\) and \(72.75\%\), respectively, while reducing overall token usage by \(21.0\%\) and \(14.3\%\) relative to Full CoT. Hard masking achieves even higher overall accuracy and TE on Qwen-32B and Llama-8B. Also, it consistently provides greater overall token savings, while the preferred variant for accuracy depends on the backbone. Overall, both strategies achieve better results across all evaluated model architectures, demonstrating the effectiveness of decoding-based intervention.

\begin{figure}[t!]
\centering
\includegraphics[width=0.92\columnwidth]{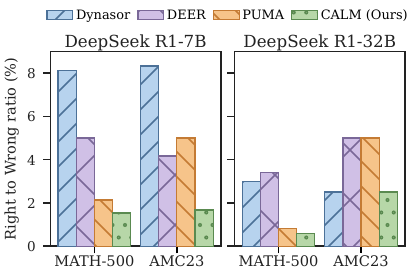}
\caption{Comparison on right-to-wrong transition rates where a problem answered correctly by Full CoT is answered incorrectly by an efficient reasoning method.}
\label{fig:rescue}
\end{figure}

\paragraph{Scales and Domains.}
The benefits of \textsc{CALM} generalize across model capacities and problem distributions. Across the DeepSeek-R1-Distill-Qwen family spanning 7B to 32B parameters, generation length is shortened while accuracy stays close to full CoT. The largest reduction is \(25.0\%\) on the 7B model. The same pattern appears on competition mathematics (MATH-500, OlympiadBench, and AIME 2025) and on GPQA-Diamond, demonstrating the applicability of our method across diverse challenging reasoning tasks.

%% file: sec/6_analysis.tex
\section{Analysis}
\label{sec:analysis}

We conduct further analysis of our method and \textsc{CALM} (Soft) is used unless otherwise noted.

\subsection{Right-to-Wrong Transitions}
\label{subsec:not_convergence}
\label{subsec:rescue}
A critical risk of reducing overthinking is that early intervention may cause errors on questions that Full CoT answers correctly. We examine the degradation rate by tracking the share of all questions on which Full CoT is correct and the method is incorrect. Figure~\ref{fig:rescue} shows the results and we find that:  
1) \textbf{Premature truncation corrupts effective reasoning.} Truncation-based early-exit baselines such as Dynasor and DEER suffer from high degradation rates. Their risk is that stopping once a local signal is met can leave out later steps that may still revise the answer. 2) \textbf{Our method Effectively Mitigates Performance Degradation.} By keeping the generation trajectory open, \textsc{CALM} substantially mitigates this corruption of performance with the lowest degradation rate.

\begin{table}[t!]
\centering
\definecolor{cotgray}{gray}{0.85}
\definecolor{calmgreen}{rgb}{0.0,0.6,0.3}
\small
\renewcommand{\arraystretch}{1.0}
\setlength{\tabcolsep}{3pt}
\adjustbox{max width=\columnwidth}{%
\begin{tabular}{lccc}
\toprule
\textbf{Configuration} / \textbf{Masked Target} & \textbf{Acc}$\uparrow$ & \textbf{Tok}$\downarrow$ & \textbf{TE}$\uparrow$ \\
\midrule
\rowcolor{cotgray!47}
Full CoT & 92.33 & 3,847 & 89.07 \\
\rowcolor{calmgreen!25}
\textsc{CALM} (\textit{Wait, Alternatively, Hmm}) & 92.27 & \textbf{2,843} & \textbf{89.77} \\
\hspace{1em}\textit{- w/o Wait} & 92.53 & 3,577 & 89.47 \\
\hspace{1em}\textit{- w/o Alternatively} & 92.00 & 3,104 & 89.30 \\
\hspace{1em}\textit{- w/o Hmm} & 92.00 & 2,896 & 89.46 \\
\midrule
\rowcolor{cotgray!47}
\multicolumn{4}{l}{\textit{Individual Suppression (reflection tokens)}} \\
\hspace{1em}\textit{Suppress Wait} & 92.20 & 3,042 & 89.55 \\
\hspace{1em}\textit{Suppress Alternatively} & 92.33 & 3,501 & 89.33 \\
\hspace{1em}\textit{Suppress Hmm} & \textbf{92.67} & 3,532 & 89.65 \\
\midrule
\rowcolor{cotgray!47}
\multicolumn{4}{l}{\textit{Individual Suppression (connective control tokens)}} \\
\hspace{1em}\textit{Suppress But} & 91.40 & 3,255 & 88.59 \\
\hspace{1em}\textit{Suppress So} & 92.40 & 3,865 & 89.13 \\
\hspace{1em}\textit{Suppress Therefore} & 91.87 & 4,000 & 88.51 \\
\hspace{1em}\textit{Suppress However} & 91.40 & 3,464 & 88.43 \\
\hspace{1em}\textit{Suppress Maybe} & 92.13 & 3,534 & 89.10 \\
\midrule
\rowcolor{cotgray!47}
\multicolumn{4}{l}{\textit{Full Vocabulary Suppression}} \\
\hspace{1em}\textit{Suppress all eight candidates} & 90.13 & 3,316 & 87.27 \\
\bottomrule
\end{tabular}%
}
\caption{ATP vocabulary ablation with \textsc{CALM} (Hard).}
\label{tab:lexicon}
\end{table}

\begin{table*}[t!]
\centering
\definecolor{tokrose}{RGB}{191,0,64}
\newcommand{\tokc}[2]{\cellcolor{tokrose!#1}#2}
\fontsize{6.5pt}{7.8pt}\selectfont
\setlength{\tabcolsep}{0.7pt}
\renewcommand{\arraystretch}{1.05}
\begin{tabular*}{\textwidth}{@{\extracolsep{\fill}}l*{13}{c}@{}}
\toprule
\multirow{2}{*}{\textbf{Variant}}
 & \multicolumn{2}{c}{\textbf{MATH-500}}
 & \multicolumn{2}{c}{\textbf{Olympiad}}
 & \multicolumn{2}{c}{\textbf{GPQA}}
 & \multicolumn{2}{c}{\textbf{AIME25}}
 & \multicolumn{2}{c}{\textbf{AMC23}}
 & \multicolumn{3}{c}{\textbf{Overall}} \\
\cmidrule(lr){2-3}\cmidrule(lr){4-5}\cmidrule(lr){6-7}\cmidrule(lr){8-9}\cmidrule(lr){10-11}\cmidrule(lr){12-14}
 & Acc$\uparrow$ & Tok$\downarrow$
 & Acc$\uparrow$ & Tok$\downarrow$
 & Acc$\uparrow$ & Tok$\downarrow$
 & Acc$\uparrow$ & Tok$\downarrow$
 & Acc$\uparrow$ & Tok$\downarrow$
 & Acc$\uparrow$ & Tok$\downarrow$ & TE$\uparrow$ \\
\midrule
Full CoT & 91.40 & 3,820 & 57.93 & 8,677 & 45.96 & 7,486 & 40.00 & 14,889 & 87.50 & 6,579 & 64.56 & 8,290 & 58.52 \\
\(\rho=0.80\) & \underline{92.00} & \tokc{22}{2,977 (-22.1\%)} & 57.04 & \tokc{21}{6,863 (-20.9\%)} & 47.98 & \tokc{34}{4,955 (-33.8\%)} & \underline{40.00} & \tokc{8}{13,641 (-8.4\%)} & \textbf{87.50} & \tokc{20}{5,247 (-20.2\%)} & \underline{64.90} & \tokc{19}{6,737 (-18.7\%)} & \underline{59.75} \\
\(\rho=0.90\) & 91.80 & \tokc{21}{3,019 (-21.0\%)} & \underline{58.37} & \tokc{23}{6,661 (-23.2\%)} & \textbf{50.00} & \tokc{32}{5,084 (-32.1\%)} & 36.67 & \tokc{4}{14,317 (-3.8\%)} & \textbf{87.50} & \tokc{20}{5,286 (-19.7\%)} & 64.87 & \tokc{17}{6,873 (-17.1\%)} & 59.64 \\
\(\rho={\text 0.95}\) & 91.00 & \tokc{17}{3,165 (-17.1\%)} & 56.89 & \tokc{19}{7,067 (-18.6\%)} & 43.43 & \tokc{26}{5,527 (-26.2\%)} & 33.33 & \tokc{5}{14,149 (-5.0\%)} & \underline{85.00} & \tokc{21}{5,165 (-21.5\%)} & 61.93 & \tokc{15}{7,015 (-15.4\%)} & 56.61 \\
\(\rho=\textbf{0.98}\) & \textbf{92.40} & \tokc{16}{3,197 (-16.3\%)} & \textbf{59.70} & \tokc{19}{7,061 (-18.6\%)} & 48.99 & \tokc{30}{5,270 (-29.6\%)} & \textbf{46.67} & \tokc{20}{11,976 (-19.6\%)} & \textbf{87.50} & \tokc{24}{4,972 (-24.4\%)} & \textbf{67.05} & \tokc{22}{6,495 (-21.7\%)} & \textbf{62.05} \\
\(\rho=0.99\) & 91.00 & \tokc{15}{3,257 (-14.7\%)} & 57.33 & \tokc{18}{7,122 (-17.9\%)} & 47.98 & \tokc{27}{5,486 (-26.7\%)} & \underline{40.00} & \tokc{14}{12,768 (-14.2\%)} & 82.50 & \tokc{12}{5,772 (-12.3\%)} & 63.76 & \tokc{17}{6,881 (-17.0\%)} & 58.52 \\
\(\rho=0.995\) & 91.40 & \tokc{13}{3,317 (-13.2\%)} & 58.22 & \tokc{15}{7,361 (-15.2\%)} & \underline{49.49} & \tokc{24}{5,709 (-23.7\%)} & \underline{40.00} & \tokc{5}{14,102 (-5.3\%)} & \underline{85.00} & \tokc{14}{5,652 (-14.1\%)} & 64.82 & \tokc{13}{7,228 (-12.8\%)} & 59.38 \\
\bottomrule
\end{tabular*}
\caption{Comparison of different soft progressive penalty factors using DeepSeek-R1-Distill-Qwen-7B .}
\label{tab:rho_overall}
\label{tab:rho}
\end{table*}

\subsection{Ablation on ATP Vocabulary}
\label{subsec:ablation}
We compare the results of suppressing different tokens in the ATP vocabulary \(\mathcal{V}_{\mathrm{ATP}}\). Results on MATH-500 by using DeepSeek-R1-Distill-Qwen-7B are shown in Table~\ref{tab:lexicon}. We first find that although suppressing all eight words in \(\mathcal{V}^{\rm full}_{\mathrm{ATP}}\) significantly reduces token length (3,316 vs. 3,847), it also decreases accuracy by 2 percentage points. Under individual suppression, both reflection and connective control tokens can almost reduce tokens, but suppressing control tokens leads to performance degradation, resulting in lower overall token efficiency, whereas suppressing anthropomorphic tokens alone maintains accuracy. We achieve the best result of 89.77 when jointly suppressing the reflection words ``\textit{Wait}'', ``\textit{Alternatively}'', and ``\textit{Hmm}''. These results illustrate the roles of different thinking tokens in the ATP vocabulary.

\subsection{Hyperparameter Sensitivity}
\label{subsec:window}
\label{subsec:rho}
We evaluate how hyperparameters balance reasoning completeness and token efficiency.

\textbf{Damping Dynamics via Soft Decay \(\rho\).}
Table~\ref{tab:rho_overall} shows the impact of different values of the progressive penalty factor \(\rho\). We observe that \(\rho=0.98\) achieves the highest accuracy, the shortest length, and the highest TE (\(62.05\)). A steeper decay (\(\rho=0.95\)) accumulates the penalty too fast, leading to an accuracy drop. Milder settings (\(\rho\ge 0.99\)) accumulate the penalty more slowly and exert weaker damping, failing to curb the generation length inflated by redundant verification.

\begin{figure}[t!]
\centering
\includegraphics[width=0.92\columnwidth]{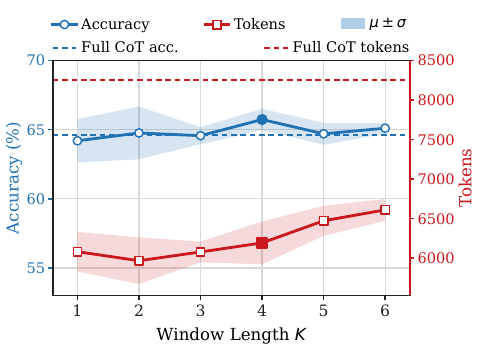}
\caption{Accuracy (blue) and tokens (red) against window length \(k\) with \textsc{CALM} (Hard). The line is the mean over three seeds.}
\label{fig:hyper}
\end{figure}

\textbf{Impact of the Consistency Window Size \(k\).}
As shown in Figure~\ref{fig:hyper}, the performance of \textsc{CALM} varies by only 1.54 percentage points across window lengths \(k\in\{1,\ldots,6\}\). The impact on tokens is slightly greater, while all settings substantially reduce token usage relative to Full CoT. We adopt \(k=4\) as the default, as it achieves the highest accuracy of 65.73\% with 6,192 tokens, providing a favorable balance between accuracy and efficiency.

\subsection{Comparison through LLM-as-a-Judge}
\label{subsec:llmjudge}
Beyond accuracy and efficiency comparisons, we evaluate reasoning-chain quality using Qwen3.5-397B-A17B through LLM-as-a-judge along dimensions of completeness, coherence, conciseness, and justification. Results in Table~\ref{tab:llmjudge} show that \textsc{CALM} achieves a higher average than the other methods and scores higher than them on completeness, coherence, and justification. In particular, after consecutive answer agreement, Full CoT still triggers ATP repeatedly, and unresolved contradictions and derivations abandoned midway make its completeness and justification lower than those of \textsc{CALM}. The detailed setup and a few case studies are given in Appendix~\ref{app:llmjudge} and Appendix~\ref{app:case_studies}, respectively.

\begin{table}[t!]
\centering
\definecolor{cotgray}{gray}{0.85}
\definecolor{calmgreen}{rgb}{0.0,0.6,0.3}
\small
\renewcommand{\arraystretch}{1.0}
\setlength{\tabcolsep}{3pt}
\adjustbox{max width=\columnwidth}{%
\begin{tabular}{lccccc}
\toprule
\textbf{Method} & \textbf{Completeness} & \textbf{Coherence} & \textbf{Conciseness} & \textbf{Justification} & \textbf{Average} \\
\midrule
\rowcolor{cotgray!47}
Full CoT & 91.0 & 90.2 & 60.5 & 89.8 & 82.9 \\
Ans.\ Conv. & 28.0 & 44.3 & \textbf{79.5} & 19.8 & 42.9 \\
Dynasor & 63.2 & 70.8 & 71.7 & 60.0 & 66.4 \\
TRACE & 81.2 & 85.0 & 61.7 & 80.0 & 77.0 \\
PUMA & 85.5 & 86.7 & 65.2 & 84.3 & 80.4 \\
DEER & 78.5 & 81.3 & 70.0 & 76.3 & 76.5 \\
\rowcolor{calmgreen!25}
\textsc{CALM} (Soft) & \textbf{94.5} & 92.3 & 62.7 & \textbf{94.0} & \textbf{85.9} \\
\rowcolor{calmgreen!25}
\textsc{CALM} (Hard) & 92.8 & \textbf{92.7} & 65.3 & 92.2 & 85.8 \\
\bottomrule
\end{tabular}%
}
\caption{Results by LLM-as-a-judge using Qwen3.5-397B-A17B. Scores run from 10 to 100 in steps of 10.}
\label{tab:llmjudge}
\end{table}

%% file: sec/7_conclusion.tex
\section{Conclusion}
\label{sec:conclusion}
We proposed consistency-aware logit masking (\textsc{CALM}), a simple yet effective training-free decoding strategy that detects consecutive answer agreement and then suppresses the logits of action transition point tokens to reduce redundant reasoning while preserving generation continuity. Extensive experiments on ten advanced reasoning models and five benchmarks show that \textsc{CALM} substantially reduces token usage while maintaining or even slightly improving accuracy relative to full CoT, offering a practical decoding-space solution to overthinking, and future work may explore adaptive masking policies, broader task domains, and combinations with reinforcement-learning or prompt-based efficiency methods.

%% file: sec/8_limitation.tex
\section*{Limitations}
\textsc{CALM} reduces generation length while keeping accuracy close to full chain-of-thought. Two limitations remain open for future work.First, the main experiments use one window length \(k\) for every item, without varying it by problem. Adapting the window length or the logit penalty to each instance, for example from token-level entropy, is a direction for future work. Second, the evaluation in this paper covers textual mathematics and science reasoning. Whether these transition tokens are used to handle external feedback in tool-use or multimodal settings is not tested here. Extending the suppression to those settings would require separating transitions that still serve a function from the meaningless exploration or verification that follows consecutive identical answers.

%% file: sec/9_appendix.tex
\raggedbottom
\renewcommand{\topfraction}{0.95}
\renewcommand{\bottomfraction}{0.7}
\renewcommand{\textfraction}{0.05}
\renewcommand{\floatpagefraction}{0.85}
\renewcommand{\dbltopfraction}{0.95}
\renewcommand{\dblfloatpagefraction}{0.85}
\setcounter{topnumber}{3}
\setcounter{dbltopnumber}{2}
\setcounter{totalnumber}{5}
\makeatletter
\setlength{\@fptop}{0pt}
\setlength{\@fpsep}{8pt plus 1fil}
\setlength{\@fpbot}{0pt plus 1fil}
\setlength{\@dblfptop}{0pt}
\setlength{\@dblfpsep}{8pt plus 1fil}
\setlength{\@dblfpbot}{0pt plus 1fil}
\makeatother

\section{Reasoning Step Segmentation}
\label{sec:step_seg}
Trial answers are probed at the level of reasoning steps, so the thought before \(\langle/\mathrm{think}\rangle\) is segmented by the deterministic, lightweight procedure of PUMA~\cite{min2026puma}.
Given the raw reasoning text, we first split the chain at blank-line boundaries, which preserves the model's natural paragraph structure.
We then assign each paragraph a coarse semantic role using simple string-level cues, such as problem setup, calculation, doubt, verification, conclusion, or general reasoning.
These labels are not used as model inputs; they only guide how adjacent paragraphs are grouped into larger reasoning steps.

The merging stage balances semantic coherence and step length.
Enumerated paragraphs or major semantic transitions are treated as natural step boundaries, while short adjacent paragraphs with compatible roles are merged when doing so preserves coherence.
We use a target step length range of $[L_{\min}, L_{\max}] = [200,1000]$ characters, and tiny trailing segments are merged into the preceding step.
This merged span is one reasoning step, and the trial answer is probed at its boundary.

\section{Trial Answer Inducing Suffix}
\label{sec:answer_suffix}
\label{sec:answer_norm}
\label{sec:probe_schedule}
For mathematics and contest problems, the answer-inducing suffix is:
\par\smallskip
\noindent\resizebox{\linewidth}{!}{\small\texttt{\textbackslash n**Final Answer**\textbackslash n\textbackslash nThe final answer is \textbackslash boxed.}}
For GPQA-Diamond, the suffix is:
\par\smallskip
\noindent\resizebox{\linewidth}{!}{\small\texttt{\textbackslash n**Final Answer**\textbackslash n\textbackslash nThe answer choice is \textbackslash boxed.}}

Before two trial answers are compared, the boxed strings are normalized by lowercasing, deleting whitespace, and stripping formatting macros.

\section{Detailed Settings in Pilot Study}
\label{sec:identical_continuation}
\paragraph{Samples.}
Figure~\ref{fig:post_agreement_dynamics} uses DeepSeek-R1-Distill-Qwen-7B on OlympiadBench . Vanilla CoT and ATP suppression use the same questions. Only questions whose trial answers are identical for four consecutive steps for the first time are included, 633 of the 675 questions. Accuracy and tokens use these 633 questions as a fixed denominator and give each question equal weight. The dashed line is the completed vanilla CoT accuracy on these questions.

\paragraph{Horizontal Axis.}
The horizontal axis is zero when the four consecutive identical trial answers end. The horizontal coordinate after that point is the mean, over the 633 questions, of the tokens added at each step after agreement. A question that has already stopped remains at the tokens it actually added after agreement. After \(\langle/\mathrm{think}\rangle\), the curve continues only once every included question has finished its thought, and that segment counts \(\langle/\mathrm{think}\rangle\) together with the following text.

\paragraph{Vertical Axis.}
At that position, accuracy is the fraction of the 633 questions whose trial answer is correct.The trial answer is produced by continuing from that prefix for at most 30 tokens and closing the answer in \texttt{\textbackslash boxed\{\}}.A finished question keeps its last correctness.

\section{ATP Surface Forms and Token Matching}
\label{app:atp_matching}

\begin{itemize}\raggedright
\item \texttt{Wait}, \texttt{wait}, \texttt{WAIT}
\item \texttt{Alternatively}, \texttt{alternatively}, \texttt{ALTERNATIVELY}
\item \texttt{H} or \texttt{h} followed by \(r\) lowercase \texttt{m}s, \(r\in\{1,\ldots,12\}\)
\end{itemize}
Each string is also included with one leading space, 60 strings in all.Each string is encoded with that model's tokenizer and \texttt{add\_special\_tokens=False}. Empty encodings are discarded, and identical token-ID sequences are deduplicated.A one-token sequence belongs to \(\mathcal{B}_t\) at every position. A longer sequence contributes its last token only when the current prefix ends with the preceding tokens. Membership in \(\mathcal{B}_t\) only marks a completion candidate. The logit changes only after agreement, by the Hard or Soft rule.

\begin{table*}[t!]
\centering
\definecolor{tokrose}{RGB}{191,0,64}
\newcommand{\tokc}[2]{\cellcolor{tokrose!#1}#2}
\fontsize{6.5pt}{7.8pt}\selectfont
\setlength{\tabcolsep}{0.7pt}
\renewcommand{\arraystretch}{1.05}
\begin{tabular*}{\textwidth}{@{\extracolsep{\fill}}l*{13}{c}@{}}
\toprule
\multirow{2}{*}{\textbf{Method}}
 & \multicolumn{2}{c}{\textbf{MATH-500}}
 & \multicolumn{2}{c}{\textbf{Olympiad}}
 & \multicolumn{2}{c}{\textbf{GPQA}}
 & \multicolumn{2}{c}{\textbf{AIME25}}
 & \multicolumn{2}{c}{\textbf{AMC23}}
 & \multicolumn{3}{c}{\textbf{Overall}} \\
\cmidrule(lr){2-3}\cmidrule(lr){4-5}\cmidrule(lr){6-7}\cmidrule(lr){8-9}\cmidrule(lr){10-11}\cmidrule(lr){12-14}
 & Acc$\uparrow$ & Tok$\downarrow$
 & Acc$\uparrow$ & Tok$\downarrow$
 & Acc$\uparrow$ & Tok$\downarrow$
 & Acc$\uparrow$ & Tok$\downarrow$
 & Acc$\uparrow$ & Tok$\downarrow$
 & Acc$\uparrow$ & Tok$\downarrow$ & TE$\uparrow$ \\
\midrule
\multicolumn{14}{c}{\textit{\textbf{Llama-3.1-Nemotron-Nano-8B}}} \\
Full CoT & 94.07 & 3,085 & 63.95 & 6,923 & 47.98 & 6,973 & 51.11 & 11,860 & 94.17 & 5,258 & 70.26 & 6,820 & 65.06 \\
Ans.\ Conv. & 57.20 & \tokc{74}{804 (-73.9\%)} & 25.28 & \tokc{80}{1,056 (-84.7\%)} & 16.67 & \tokc{80}{336 (-95.2\%)} & 4.44 & \tokc{80}{1,183 (-90.0\%)} & 28.33 & \tokc{80}{807 (-84.7\%)} & 26.39 & \tokc{80}{837 (-87.7\%)} & 25.59 \\
Dynasor & 93.00 & 3,288 (+6.6\%) & \underline{63.95} & 7,081 (+2.3\%) & 46.97 & 7,664 (+9.9\%) & 45.56 & 12,651 (+6.7\%) & \underline{93.33} & 5,449 (+3.6\%) & 68.56 & 7,227 (+6.0\%) & 63.12 \\
TRACE & 90.67 & \tokc{26}{2,296 (-25.6\%)} & 59.16 & \tokc{46}{3,734 (-46.1\%)} & 37.21 & \tokc{67}{2,313 (-66.8\%)} & 35.56 & \tokc{42}{6,814 (-42.5\%)} & 85.83 & \tokc{33}{3,500 (-33.4\%)} & 61.69 & \tokc{45}{3,731 (-45.3\%)} & 58.52 \\
PUMA & 92.73 & \tokc{18}{2,537 (-17.8\%)} & 62.96 & \tokc{27}{5,023 (-27.4\%)} & 46.63 & \tokc{15}{5,911 (-15.2\%)} & 46.67 & \tokc{16}{10,012 (-15.6\%)} & 91.67 & \tokc{17}{4,370 (-16.9\%)} & 68.13 & \tokc{18}{5,571 (-18.3\%)} & 63.70 \\
DEER & 92.13 & \tokc{29}{2,202 (-28.6\%)} & 63.56 & \tokc{23}{5,355 (-22.6\%)} & \underline{48.82} & 7,303 (+4.7\%) & \underline{50.00} & \tokc{5}{11,235 (-5.3\%)} & 91.67 & \tokc{15}{4,453 (-15.3\%)} & 69.24 & \tokc{10}{6,110 (-10.4\%)} & 64.47 \\
\textsc{CALM} (Soft) & \underline{93.33} & \tokc{10}{2,770 (-10.2\%)} & \textbf{64.20} & \tokc{10}{6,262 (-9.5\%)} & \textbf{50.34} & \tokc{12}{6,120 (-12.2\%)} & \underline{50.00} & \tokc{11}{10,537 (-11.2\%)} & \textbf{94.17} & \tokc{12}{4,643 (-11.7\%)} & \textbf{70.41} & \tokc{11}{6,066 (-11.1\%)} & \textbf{65.67} \\
\textsc{CALM} (Hard) & \textbf{93.80} & \tokc{14}{2,646 (-14.2\%)} & 63.36 & \tokc{15}{5,913 (-14.6\%)} & 46.97 & \tokc{17}{5,757 (-17.4\%)} & \textbf{51.11} & \tokc{15}{10,097 (-14.9\%)} & \underline{93.33} & \tokc{14}{4,509 (-14.2\%)} & \underline{69.71} & \tokc{15}{5,784 (-15.2\%)} & \underline{65.15} \\
\midrule
\multicolumn{14}{c}{\textit{\textbf{Qwen3-4B}}} \\
Full CoT & 94.07 & 5,042 & 66.96 & 10,492 & 55.39 & 8,077 & 68.89 & 17,826 & 96.67 & 7,862 & 76.40 & 9,860 & 69.54 \\
Ans.\ Conv. & 57.07 & \tokc{80}{976 (-80.6\%)} & 27.46 & \tokc{80}{1,314 (-87.5\%)} & 37.88 & \tokc{80}{322 (-96.0\%)} & 8.89 & \tokc{80}{1,938 (-89.1\%)} & 37.50 & \tokc{80}{1,098 (-86.0\%)} & 33.76 & \tokc{80}{1,130 (-88.5\%)} & 32.69 \\
Dynasor & 93.13 & 6,009 (+19.2\%) & 65.38 & 11,966 (+14.0\%) & \textbf{55.22} & 8,748 (+8.3\%) & 63.33 & 19,919 (+11.7\%) & 95.00 & 9,410 (+19.7\%) & 74.41 & 11,210 (+13.7\%) & 66.89 \\
TRACE & 84.80 & \tokc{47}{2,679 (-46.9\%)} & 51.21 & \tokc{66}{3,557 (-66.1\%)} & 41.58 & \tokc{75}{1,983 (-75.4\%)} & 34.44 & \tokc{69}{5,583 (-68.7\%)} & 77.50 & \tokc{50}{3,957 (-49.7\%)} & 57.91 & \tokc{64}{3,552 (-64.0\%)} & 54.87 \\
PUMA & 91.73 & \tokc{38}{3,103 (-38.5\%)} & 62.32 & \tokc{51}{5,147 (-50.9\%)} & 53.87 & \tokc{33}{5,407 (-33.1\%)} & 56.67 & \tokc{18}{14,647 (-17.8\%)} & 93.33 & \tokc{31}{5,432 (-30.9\%)} & 71.59 & \tokc{32}{6,747 (-31.6\%)} & 66.43 \\
DEER & 93.13 & \tokc{22}{3,922 (-22.2\%)} & \textbf{66.72} & \tokc{10}{9,466 (-9.8\%)} & 53.70 & \tokc{7}{7,542 (-6.6\%)} & 62.22 & \tokc{7}{16,603 (-6.9\%)} & 95.83 & \tokc{13}{6,856 (-12.8\%)} & 74.32 & \tokc{10}{8,878 (-10.0\%)} & 67.97 \\
\textsc{CALM} (Soft) & \textbf{94.27} & \tokc{16}{4,247 (-15.8\%)} & \underline{66.42} & \tokc{14}{9,039 (-13.8\%)} & 53.37 & \tokc{16}{6,763 (-16.3\%)} & \textbf{67.78} & \tokc{12}{15,725 (-11.8\%)} & \underline{96.67} & \tokc{16}{6,572 (-16.4\%)} & \textbf{75.70} & \tokc{14}{8,469 (-14.1\%)} & \textbf{69.56} \\
\textsc{CALM} (Hard) & \underline{93.67} & \tokc{20}{4,040 (-19.9\%)} & 65.98 & \tokc{17}{8,742 (-16.7\%)} & \underline{54.21} & \tokc{19}{6,570 (-18.7\%)} & \underline{64.44} & \tokc{11}{15,896 (-10.8\%)} & \textbf{97.50} & \tokc{23}{6,065 (-22.9\%)} & \underline{75.16} & \tokc{16}{8,263 (-16.2\%)} & \underline{69.14} \\
\midrule
\multicolumn{14}{c}{\textit{\textbf{Qwen3-8B}}} \\
Full CoT & 94.27 & 5,242 & 68.15 & 11,197 & 58.59 & 8,806 & 68.89 & 17,746 & 96.67 & 8,543 & 77.31 & 10,307 & 70.23 \\
Ans.\ Conv. & 59.80 & \tokc{80}{860 (-83.6\%)} & 27.80 & \tokc{80}{1,153 (-89.7\%)} & 40.91 & \tokc{80}{309 (-96.5\%)} & 10.00 & \tokc{80}{1,965 (-88.9\%)} & 45.00 & \tokc{80}{1,143 (-86.6\%)} & 36.70 & \tokc{80}{1,086 (-89.5\%)} & 35.67 \\
Dynasor & 92.67 & 5,654 (+7.9\%) & 66.67 & 11,684 (+4.3\%) & 54.04 & \tokc{56}{3,875 (-56.0\%)} & \textbf{73.33} & 20,179 (+13.7\%) & 94.17 & 9,272 (+8.5\%) & 76.17 & \tokc{2}{10,133 (-1.7\%)} & 69.17 \\
TRACE & 86.13 & \tokc{45}{2,881 (-45.0\%)} & 51.51 & \tokc{65}{3,883 (-65.3\%)} & 52.53 & \tokc{75}{2,223 (-74.8\%)} & 34.44 & \tokc{75}{4,480 (-74.8\%)} & 76.67 & \tokc{53}{3,992 (-53.3\%)} & 60.26 & \tokc{66}{3,492 (-66.1\%)} & 57.26 \\
PUMA & 92.27 & \tokc{41}{3,091 (-41.0\%)} & 61.78 & \tokc{52}{5,403 (-51.7\%)} & 55.89 & \tokc{40}{5,275 (-40.1\%)} & 62.22 & \tokc{20}{14,138 (-20.3\%)} & 90.00 & \tokc{40}{5,118 (-40.1\%)} & 72.43 & \tokc{36}{6,605 (-35.9\%)} & 67.36 \\
DEER & 94.07 & \tokc{34}{3,435 (-34.5\%)} & 66.52 & \tokc{16}{9,386 (-16.2\%)} & \underline{57.74} & \tokc{2}{8,604 (-2.3\%)} & \underline{71.11} & \tokc{12}{15,646 (-11.8\%)} & \textbf{97.50} & \tokc{16}{7,165 (-16.1\%)} & \underline{77.39} & \tokc{14}{8,847 (-14.2\%)} & \underline{71.05} \\
\textsc{CALM} (Soft) & \underline{94.27} & \tokc{17}{4,325 (-17.5\%)} & \textbf{67.80} & \tokc{15}{9,566 (-14.6\%)} & \textbf{59.26} & \tokc{18}{7,231 (-17.9\%)} & \underline{71.11} & \tokc{7}{16,568 (-6.6\%)} & \underline{95.00} & \tokc{16}{7,188 (-15.9\%)} & \textbf{77.49} & \tokc{13}{8,976 (-12.9\%)} & \textbf{71.08} \\
\textsc{CALM} (Hard) & \textbf{94.53} & \tokc{23}{4,020 (-23.3\%)} & \underline{67.46} & \tokc{18}{9,160 (-18.2\%)} & 57.58 & \tokc{23}{6,759 (-23.2\%)} & 66.67 & \tokc{13}{15,471 (-12.8\%)} & \textbf{97.50} & \tokc{19}{6,920 (-19.0\%)} & 76.75 & \tokc{18}{8,466 (-17.9\%)} & 70.62 \\
\midrule
\multicolumn{14}{c}{\textit{\textbf{Qwen3-30B-A3B}}} \\
Full CoT & 95.93 & 5,176 & 76.25 & 12,520 & 75.59 & 7,441 & 83.33 & 17,308 & 100.00 & 8,708 & 86.22 & 10,231 & 79.17 \\
Ans.\ Conv. & 59.80 & \tokc{80}{666 (-87.1\%)} & 23.85 & \tokc{80}{1,105 (-91.2\%)} & 34.85 & \tokc{80}{304 (-95.9\%)} & 0.00 & \tokc{80}{982 (-94.3\%)} & 29.17 & \tokc{80}{870 (-90.0\%)} & 29.53 & \tokc{80}{785 (-92.3\%)} & 28.77 \\
Dynasor & 94.33 & \tokc{5}{4,901 (-5.3\%)} & 72.59 & \tokc{2}{12,272 (-2.0\%)} & 60.61 & \tokc{50}{3,690 (-50.4\%)} & 78.89 & \tokc{3}{16,779 (-3.1\%)} & 98.33 & \tokc{2}{8,554 (-1.8\%)} & 80.95 & \tokc{10}{9,239 (-9.7\%)} & 74.41 \\
TRACE & 92.67 & \tokc{24}{3,912 (-24.4\%)} & 66.52 & \tokc{44}{6,985 (-44.2\%)} & 57.74 & \tokc{70}{2,237 (-69.9\%)} & 56.67 & \tokc{34}{11,476 (-33.7\%)} & 89.17 & \tokc{23}{6,713 (-22.9\%)} & 72.55 & \tokc{39}{6,265 (-38.8\%)} & 67.69 \\
PUMA & \underline{95.33} & \tokc{30}{3,623 (-30.0\%)} & 73.29 & \tokc{33}{8,396 (-32.9\%)} & 73.90 & \tokc{17}{6,213 (-16.5\%)} & 78.89 & \tokc{13}{15,020 (-13.2\%)} & \textbf{100.00} & \tokc{27}{6,395 (-26.6\%)} & 84.28 & \tokc{23}{7,929 (-22.5\%)} & 78.44 \\
DEER & \textbf{96.00} & \tokc{22}{4,057 (-21.6\%)} & \textbf{76.74} & \tokc{10}{11,285 (-9.9\%)} & \textbf{75.08} & 7,695 (+3.4\%) & \underline{82.22} & 17,291 (-0.1\%) & \underline{99.17} & \tokc{6}{8,187 (-6.0\%)} & 85.84 & \tokc{5}{9,703 (-5.2\%)} & 79.06 \\
\textsc{CALM} (Soft) & \textbf{96.00} & \tokc{11}{4,588 (-11.4\%)} & \textbf{76.74} & \tokc{11}{11,190 (-10.6\%)} & \underline{74.75} & \tokc{20}{5,955 (-20.0\%)} & \textbf{83.33} & \tokc{5}{16,422 (-5.1\%)} & \textbf{100.00} & \tokc{11}{7,750 (-11.0\%)} & \textbf{86.16} & \tokc{10}{9,181 (-10.3\%)} & \textbf{79.65} \\
\textsc{CALM} (Hard) & \textbf{96.00} & \tokc{18}{4,237 (-18.1\%)} & \underline{76.35} & \tokc{14}{10,770 (-14.0\%)} & 73.74 & \tokc{26}{5,475 (-26.4\%)} & \textbf{83.33} & \tokc{9}{15,709 (-9.2\%)} & \textbf{100.00} & \tokc{18}{7,141 (-18.0\%)} & \underline{85.88} & \tokc{15}{8,666 (-15.3\%)} & \underline{79.64} \\
\midrule
\multicolumn{14}{c}{\textit{\textbf{Qwen3-32B}}} \\
Full CoT & 95.13 & 4,509 & 69.53 & 9,763 & 69.70 & 6,733 & 70.00 & 15,724 & 99.17 & 6,526 & 80.71 & 8,651 & 74.48 \\
Ans.\ Conv. & 70.87 & \tokc{78}{1,002 (-77.8\%)} & 33.73 & \tokc{80}{1,385 (-85.8\%)} & 47.47 & \tokc{80}{314 (-95.3\%)} & 23.33 & \tokc{80}{3,175 (-79.8\%)} & 40.00 & \tokc{79}{1,380 (-78.9\%)} & 43.08 & \tokc{80}{1,451 (-83.2\%)} & 41.73 \\
Dynasor & 93.87 & 4,842 (+7.4\%) & 69.19 & 10,545 (+8.0\%) & 57.74 & \tokc{52}{3,234 (-52.0\%)} & 67.78 & 18,404 (+17.0\%) & \textbf{99.17} & 7,025 (+7.6\%) & 77.55 & 8,810 (+1.8\%) & 71.23 \\
TRACE & 91.20 & \tokc{30}{3,155 (-30.0\%)} & 60.35 & \tokc{50}{4,935 (-49.5\%)} & 62.46 & \tokc{56}{2,948 (-56.2\%)} & 44.44 & \tokc{62}{5,975 (-62.0\%)} & 85.83 & \tokc{34}{4,321 (-33.8\%)} & 68.86 & \tokc{51}{4,267 (-50.7\%)} & 65.31 \\
PUMA & 93.33 & \tokc{32}{3,046 (-32.4\%)} & 65.93 & \tokc{44}{5,440 (-44.3\%)} & 67.17 & \tokc{37}{4,252 (-36.8\%)} & 63.33 & \tokc{30}{11,060 (-29.7\%)} & 90.83 & \tokc{31}{4,475 (-31.4\%)} & 76.12 & \tokc{35}{5,655 (-34.6\%)} & 71.64 \\
DEER & 94.13 & \tokc{25}{3,385 (-24.9\%)} & 68.89 & \tokc{12}{8,595 (-12.0\%)} & 67.51 & \tokc{2}{6,628 (-1.6\%)} & \textbf{75.56} & \tokc{4}{15,034 (-4.4\%)} & 96.67 & \tokc{17}{5,433 (-16.7\%)} & \underline{80.55} & \tokc{10}{7,815 (-9.7\%)} & \underline{74.78} \\
\textsc{CALM} (Soft) & \textbf{95.60} & \tokc{17}{3,739 (-17.1\%)} & \underline{69.28} & \tokc{11}{8,647 (-11.4\%)} & \underline{68.86} & \tokc{16}{5,646 (-16.1\%)} & 70.00 & \tokc{6}{14,807 (-5.8\%)} & \underline{97.50} & \tokc{16}{5,512 (-15.5\%)} & 80.25 & \tokc{11}{7,670 (-11.3\%)} & 74.56 \\
\textsc{CALM} (Hard) & \underline{95.00} & \tokc{19}{3,633 (-19.4\%)} & \textbf{69.53} & \tokc{14}{8,391 (-14.1\%)} & \textbf{70.71} & \tokc{19}{5,445 (-19.1\%)} & \underline{72.22} & \tokc{9}{14,385 (-8.5\%)} & \textbf{99.17} & \tokc{18}{5,351 (-18.0\%)} & \textbf{81.33} & \tokc{14}{7,441 (-14.0\%)} & \textbf{75.77} \\
\midrule
\multicolumn{14}{c}{\textit{\textbf{QwQ-32B}}} \\
Full CoT & 95.07 & 4,330 & 71.16 & 9,612 & 67.85 & 7,796 & 70.00 & 16,263 & 99.17 & 7,550 & 80.65 & 9,110 & 74.17 \\
Ans.\ Conv. & 72.53 & \tokc{74}{1,129 (-73.9\%)} & 34.77 & \tokc{80}{1,709 (-82.2\%)} & 36.87 & \tokc{80}{344 (-95.6\%)} & 14.44 & \tokc{72}{4,488 (-72.4\%)} & 56.67 & \tokc{73}{2,039 (-73.0\%)} & 43.06 & \tokc{79}{1,942 (-78.7\%)} & 41.29 \\
Dynasor & 94.40 & 5,420 (+25.2\%) & 69.73 & 11,326 (+17.8\%) & 58.42 & \tokc{45}{4,324 (-44.5\%)} & \textbf{72.22} & 20,520 (+26.2\%) & \textbf{98.33} & 9,628 (+27.5\%) & 78.62 & 10,244 (+12.4\%) & 71.57 \\
TRACE & 90.87 & \tokc{34}{2,878 (-33.5\%)} & 60.59 & \tokc{53}{4,556 (-52.6\%)} & 56.40 & \tokc{70}{2,331 (-70.1\%)} & 48.89 & \tokc{43}{9,317 (-42.7\%)} & 83.33 & \tokc{39}{4,589 (-39.2\%)} & 68.02 & \tokc{48}{4,734 (-48.0\%)} & 64.14 \\
PUMA & 90.47 & \tokc{25}{3,255 (-24.8\%)} & 61.53 & \tokc{31}{6,586 (-31.5\%)} & 62.80 & \tokc{27}{5,714 (-26.7\%)} & 62.22 & \tokc{23}{12,554 (-22.8\%)} & 88.33 & \tokc{25}{5,689 (-24.6\%)} & 73.07 & \tokc{26}{6,760 (-25.8\%)} & 67.91 \\
DEER & \underline{95.47} & \tokc{17}{3,592 (-17.0\%)} & \textbf{70.72} & \tokc{10}{8,604 (-10.5\%)} & 65.15 & 7,975 (+2.3\%) & 64.44 & \tokc{7}{15,062 (-7.4\%)} & 96.67 & \tokc{8}{6,931 (-8.2\%)} & 78.49 & \tokc{7}{8,433 (-7.4\%)} & 72.37 \\
\textsc{CALM} (Soft) & \textbf{95.53} & \tokc{19}{3,517 (-18.8\%)} & \underline{70.62} & \tokc{18}{7,923 (-17.6\%)} & \underline{66.67} & \tokc{17}{6,439 (-17.4\%)} & \underline{70.00} & \tokc{13}{14,125 (-13.1\%)} & 95.83 & \tokc{21}{5,966 (-21.0\%)} & \underline{79.73} & \tokc{17}{7,594 (-16.6\%)} & \underline{74.08} \\
\textsc{CALM} (Hard) & 95.20 & \tokc{24}{3,278 (-24.3\%)} & 68.94 & \tokc{21}{7,567 (-21.3\%)} & \textbf{67.34} & \tokc{26}{5,760 (-26.1\%)} & \underline{70.00} & \tokc{13}{14,163 (-12.9\%)} & \underline{97.50} & \tokc{18}{6,174 (-18.2\%)} & \textbf{79.80} & \tokc{19}{7,388 (-18.9\%)} & \textbf{74.27} \\
\bottomrule
\end{tabular*}
\caption{Averaged results over three seeds across additional reasoning models. Acc: accuracy,  Tok: token count, TE: token efficiency. Values in parentheses are relative changes in tokens from Full CoT. The best and second-best accuracy and TE values among the efficient-reasoning methods are shown in \textbf{bold} and \underline{underlined}, respectively; Full CoT is excluded from this ranking.}
\label{tab:more_models}
\end{table*}

\section{Main Results on Additional Models}
\label{app:more_models}

Table~\ref{tab:more_models} uses the same columns as Table~\ref{tab:main} and reports accuracy and tokens on the five benchmarks for Llama-3.1-Nemotron-Nano-8B, Qwen3 (4B, 8B, 30B-A3B, and 32B), and QwQ-32B, together with the equal-weight overall and its TE.
Each cell is the mean over three seeds, and the parentheses on tokens are the change relative to Full CoT for the same model and benchmark.

\section{Experimental Details}
\label{sec:instruction_decoding}

For AMC 2023, we use the 40 questions from LIMO~\cite{ye2025limo}, following DEER.

All methods run under one vLLM host~\cite{kwon2023pagedattention}.The main generation budget is \(32{,}768\) tokens.If this budget ends before the thought closes, the answer continuation is at most \(2{,}048\) tokens.The prompt reserve is \(3{,}072\) tokens, and \texttt{max\allowbreak\_model\allowbreak\_len} is \(37{,}888\).For a question with a finite agreement boundary \(b^{\star}\), the generated thought is kept.When the model emits \texttt{</think>} within this budget, the following suffix is appended:
\begin{quote}
\small
\texttt{The final answer is \textbackslash boxed}
\end{quote}
If this budget ends before that marker, generation continues with the following suffix:
\begin{quote}
\small
\texttt{\textbackslash n</think>\textbackslash n\textbackslash nThe final answer is \textbackslash boxed}
\end{quote}
Delivered reasoning, the final answer, and trial answers use the models' recommended reasoning settings, with temperature 0.6 and top-\(p\) 0.95.The instruction on mathematics and contest benchmarks is:
\begin{quote}
\small
Solve the following math problem. Please reason step by step, and put your final answer within \texttt{\textbackslash boxed\{\}}.
\end{quote}
The instruction on GPQA-Diamond is:
\begin{quote}
\small
Answer the following multiple choice question. Please reason step by step, and put your choice letter (A, B, C, or D) without any other text within \texttt{\textbackslash boxed\{\}} in the end.
\end{quote}
These instructions and the answer-inducing suffixes follow the default prompts of PUMA~\cite{min2026puma}.

\begin{table*}[t!]
\centering
\footnotesize
\setlength{\tabcolsep}{3pt}
\renewcommand{\arraystretch}{1.05}
\begin{tabular*}{\textwidth}{@{\extracolsep{\fill}}llccccc@{}}
\toprule
\textbf{Model} & \textbf{Benchmark} & \textbf{CALM} & \textbf{Ans.\ Conv.} & \textbf{Dynasor} & \textbf{PUMA} & \textbf{DEER} \\
\midrule
\multirow{5}{*}{DeepSeek-R1-Distill-Qwen-7B}
 & MATH-500 (500) & 421.7 & 458.3 & 352.0 & 307.0 & 397.3 \\
 & Olympiad (675) & 634.0 & 649.3 & 520.7 & 464.0 & 420.3 \\
 & GPQA (198) & 180.7 & 198.0 & 182.3 & 49.7 & 0.0 \\
 & AIME25 (30) & 28.0 & 30.0 & 27.0 & 11.7 & 9.7 \\
 & AMC23 (40) & 36.7 & 39.3 & 33.7 & 22.3 & 27.3 \\
\midrule
\multirow{5}{*}{DeepSeek-R1-Distill-Qwen-14B}
 & MATH-500 (500) & 414.7 & 483.3 & 223.0 & 288.3 & 420.7 \\
 & Olympiad (675) & 641.7 & 665.7 & 381.7 & 441.3 & 432.0 \\
 & GPQA (198) & 180.3 & 197.7 & 170.0 & 86.0 & 0.0 \\
 & AIME25 (30) & 27.3 & 29.0 & 22.3 & 11.0 & 8.3 \\
 & AMC23 (40) & 34.7 & 38.0 & 21.7 & 20.0 & 27.0 \\
\midrule
\multirow{5}{*}{DeepSeek-R1-Distill-Qwen-32B}
 & MATH-500 (500) & 266.0 & 306.7 & 187.0 & 158.0 & 242.0 \\
 & Olympiad (675) & 544.7 & 570.0 & 401.0 & 348.0 & 345.7 \\
 & GPQA (198) & 176.7 & 198.0 & 171.7 & 64.0 & 0.0 \\
 & AIME25 (30) & 26.3 & 29.7 & 23.0 & 13.0 & 9 \\
 & AMC23 (40) & 30.7 & 36.7 & 20.3 & 15.7 & 28 \\
\midrule
\multirow{5}{*}{DeepSeek-R1-Distill-Llama-8B}
 & MATH-500 (500) & 321.3 & 220.3 & 246.7 & 218.7 & 224.0 \\
 & Olympiad (675) & 596.3 & 548.3 & 460.0 & 409.3 & 263.0 \\
 & GPQA (198) & 181.3 & 198.0 & 182.3 & 53.7 & 0.0 \\
 & AIME25 (30) & 27.3 & 30.0 & 24.7 & 8.3 & 5.0 \\
 & AMC23 (40) & 32.7 & 29.0 & 26.0 & 19.3 & 16.3 \\
\bottomrule
\end{tabular*}
\caption{Questions stopped, or intervened on by \textsc{CALM}, as the mean over three seeds. The parenthetical value is the number of questions. \textsc{CALM} counts four consecutive identical trial answers. The other columns count questions on which that method's stopping rule fires.}
\label{tab:stop_coverage}
\end{table*}

\section{Stop Coverage on the Main-Table Cells}
\label{app:stop_coverage}
Table~\ref{tab:stop_coverage} reports, for the four main-table models and five benchmarks, how many questions \textsc{CALM} intervenes on and how many the early-exit methods stop, as the mean over three seeds.
The parenthetical value is the number of questions on that benchmark.
\textsc{CALM} counts questions with four consecutive identical trial answers.
The other columns count questions on which that method's stopping rule fires.
In most cells,Answer Convergence stops on more questions than \textsc{CALM} intervenes on.
DEER's confidence exits are fewer, except on DeepSeek-R1-Distill-Qwen-14B MATH-500.

\section{Additional Right-to-Wrong Rates}
\label{app:rescue_more}

\begin{figure*}[t]
\centering
\includegraphics[width=\textwidth]{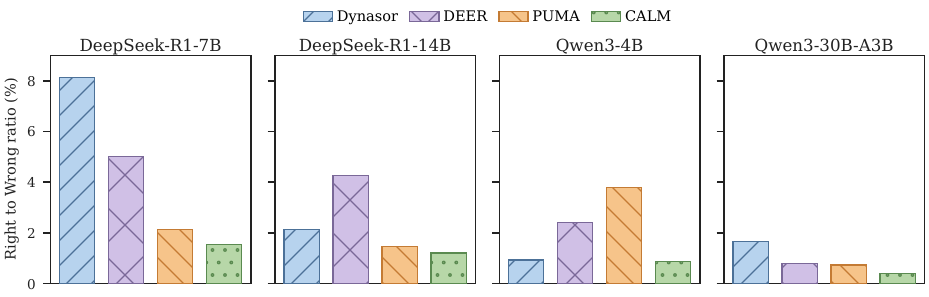}

\caption{Right-to-wrong rates on MATH-500. Panels pool three seeds and show DeepSeek-R1-7B, DeepSeek-R1-14B, Qwen3-4B, and Qwen3-30B-A3B from left to right. A right-to-wrong transition occurs when full CoT answers a problem correctly and the method answers it incorrectly. \textsc{CALM} is the lowest of the four methods on each panel.}
\label{fig:rescue_more}
\end{figure*}

Figure~\ref{fig:rescue_more} extends Figure~\ref{fig:rescue} to four  models on MATH-500. The rate is the share of questions that full CoT answers correctly and the method answers incorrectly. \textsc{CALM} Soft is the lowest among Dynasor, DEER, PUMA, and \textsc{CALM} on every panel. These additional results demonstrate the stability of our method, whereas other truncation-based methods are indeed prone to depriving the model of opportunities for subsequent correction.

\tcbset{
  calmbox/.style={
    width=\linewidth,
    colback=black!3,
    colframe=black!45,
    boxrule=0.4pt,
    arc=1pt,
    breakable,
  },
}

\section{Settings for LLM-as-a-Judge}
\label{app:llmjudge}

We evaluate the quality of retained reasoning chains with an LLM judge.
The goal is to measure whether the retained chain remains a readable and sufficient derivation, beyond final-answer accuracy.

\paragraph{Judge Model and Protocol.}
The judge is Qwen3.5-397B-A17B, called at temperature 0 with at most 1,024 output tokens. Each call receives the problem, the retained reasoning, and the answer submitted by that method. The judge writes a one-sentence rationale before the scores. 

\paragraph{Scope.}
The evaluated chains are generations of DeepSeek-R1-Distill-Qwen-7B on MATH-500.
We randomly draw 60 questions, and every method is scored on these 60 questions.
The model's input is the problem, the retained reasoning, and the submitted answer.
When a method stops early, the retained reasoning ends at that stop; otherwise it is the full reasoning.

\paragraph{Rubric.}
Each chain is scored from 10 to 100 in steps of 10 on completeness, coherence, conciseness, and justification, following the four dimensions of PUMA~\cite{min2026puma}.
Completeness asks whether the essential steps from the problem to the stated answer are present.
Coherence asks whether the progression is connected and whether revisions are resolved.
Conciseness asks whether repetition or unproductive detours are present, and missing steps are scored under completeness and justification instead.
Justification asks whether the written derivation supports the stated answer.
The prompt states the anchors at 100, 70, 40, and 10.

\begin{tcolorbox}[calmbox]
\textbf{Role.}
You evaluate the quality of a written mathematical derivation.
You receive the problem, the retained reasoning, and the stated answer.
The stated answer is a claim to assess, not a reference answer.

\medskip
\textbf{Boundary.}
Do not solve the problem independently, and do not grade the chain against your own answer or an external reference.
Do inspect violations of the problem's conditions, invalid arithmetic or logical transitions, missing essential steps, and unresolved contradictions.
Ground every deduction in the supplied text.
Do not invent a missing argument, and do not treat the stated answer itself as evidence for the missing steps.
Ordinary algebraic compression is acceptable.
A short chain can be sufficient, and a long chain can remain incomplete.
Score the four dimensions separately: an unsupported answer can still be concise.

\medskip
\textbf{Scores.}
Use only 10, 20, 30, 40, 50, 60, 70, 80, 90, and 100.
Completeness asks whether the essential reasoning from the problem to the stated answer is present: 100 if the essential links and conditions are covered, 70 if the main route is developed but some support is thin, 40 if a major part is missing, and 10 if the answer is asserted without a derivation.
Coherence asks whether one connected progression can be followed: 100 if it is clear and revisions are resolved, 70 if it is mostly followable, 40 if disconnection or unresolved conflict disrupts the route, and 10 if no followable line can be identified.
Conciseness asks how much unnecessary material appears: 100 if there is no material repetition or unproductive detour, 70 if some avoidable repetition is present, 40 if redundancy is substantial, and 10 if repetition dominates.
Do not lower conciseness merely because essential reasoning is missing.
Justification asks whether the text explains why this stated answer follows: 100 if it is fully supported, 70 if an important explanation remains implicit, 40 if the decisive connection is missing or defective, and 10 if the answer is unsupported or contradicted.

\medskip
\textbf{Output.}
Return only one JSON object, with a one-sentence rationale first.
{\raggedright\ttfamily \{"brief\_rationale": "...", "completeness": 10, "coherence": 10, "conciseness": 10, "justification": 10\}\par}
The numbers illustrate the format. Choose the actual scores from the permitted scale.
\end{tcolorbox}

\tcbset{
  casebase/.style={
    enhanced,
    boxrule=0.45pt,
    arc=1.5pt,
    left=5pt, right=5pt, top=4pt, bottom=4pt,
    fontupper=\small,
    colframe=black!40,
  },
  casegray/.style={casebase, colback=black!8, colframe=black!40},
  casegreen/.style={casebase, colback=green!12, colframe=green!50!black},
  caseans/.style={casebase, colback=yellow!30, colframe=yellow!65!black},
  caseblue/.style={casebase, colback=blue!10, colframe=blue!45!black},
  caseouter/.style={
    enhanced,
    colback=white,
    colframe=black!65,
    boxrule=0.8pt,
    arc=2pt,
    left=8pt, right=8pt, top=8pt, bottom=8pt,
    fonttitle=\normalsize,
    coltitle=black,
    colbacktitle=black!4,
    toptitle=4pt,
    bottomtitle=4pt,
  },
}
\newcommand{\deertrial}[2]{%
\begin{tcolorbox}[caseans]
\small The final answer is \(\boxed{#1}\)\ \(\longrightarrow\)\ confidence #2%
\end{tcolorbox}}

\section{Case Study}
\label{app:case_studies}
The following traces are generations of DeepSeek-R1-Distill-Qwen-7B on MATH-500.

\begin{figure*}[p]
\centering
\begin{tcolorbox}[caseouter, title={Question 148. Positive reals satisfy \(xyz=2\). Find the minimum of \(x^4+4y^2+4z^4\). Ground truth: 16.}]
\begin{minipage}[t]{0.48\linewidth}
\centering\textbf{\textsc{CALM}}\par\vspace{0.45em}
\begin{tcolorbox}[casegray]
\textbf{Alternatively}, maybe I can use substitution. Since \(xyz=2\), I can express one variable in terms of the others. For example, \(z=\frac{2}{xy}\).
\end{tcolorbox}
\vspace{0.55em}
\begin{tcolorbox}[casegray]
Then, substitute this into the expression \(x^4+4y^2+4z^4\).

Let me try that:
\end{tcolorbox}
\vspace{0.55em}
\begin{tcolorbox}[casegreen]
substitute \(z=\frac{2}{xy}\), so \(z^4=\left(\frac{2}{xy}\right)^4=\frac{16}{x^4 y^4}\).

Therefore, the expression becomes \(x^4+4y^2+\frac{64}{x^4 y^4}\).
\end{tcolorbox}
\vspace{0.55em}
\begin{tcolorbox}[casegreen]
\(x=\sqrt{2}\), \(y=\sqrt{2}\), and \(z=1\).

\(x^4=4\), \(4y^2=8\), \(4z^4=4\). Adding them up: \(4+8+4=16\).

So, the minimum value is 16.
\end{tcolorbox}
\vspace{0.55em}
\begin{tcolorbox}[caseblue]
\texttt{</think>} The final answer is \(\boxed{16}\).
\end{tcolorbox}
\vspace{0.45em}
{\normalsize\bfseries Tokens: 4{,}297\quad \boxed{\boldsymbol{16}}}\ \cmark
\end{minipage}
\hfill
\begin{minipage}[t]{0.48\linewidth}
\centering\textbf{Full CoT}\par\vspace{0.45em}
\begin{tcolorbox}[casegray]
\textbf{Alternatively}, maybe I can use substitution. Since \(xyz=2\), I can express one variable in terms of the others. For example, \(z=\frac{2}{xy}\).
\end{tcolorbox}
\vspace{0.55em}
\begin{tcolorbox}[casegray]
Then, substitute this into the expression \(x^4+4y^2+4z^4\).

Let me try that:
\end{tcolorbox}
\vspace{0.55em}
\begin{tcolorbox}[casegray]
Substituting \(z=\frac{2}{xy}\), we get \(x^4+4y^2+\frac{64}{x^4 y^4}\).
\end{tcolorbox}
\vspace{0.55em}
\begin{tcolorbox}[casegray]
\textbf{Hmm}, now we have an expression in terms of \(x\) and \(y\). Maybe I can set partial derivatives with respect to \(x\) and \(y\) to zero to find the minimum. Let me try calculus.
\end{tcolorbox}
\vspace{0.55em}
\begin{tcolorbox}[casegray]
Set both partial derivatives equal to zero:
\[
4x^3-\frac{256}{x^5 y^4}=0,\qquad
8y-\frac{256}{x^4 y^5}=0.
\]
From the first equation, \(x^2 y=2\).
\end{tcolorbox}
\vspace{0.55em}
\begin{tcolorbox}[casegray]
\textbf{Wait}, that can't be right because \(1+8+4\) is 13, so \(13\sqrt{2}\). \textbf{Hmm}, but let me verify the calculations again.
\end{tcolorbox}
\vspace{0.55em}
\begin{tcolorbox}[casegray]
\textbf{Wait}, maybe I made a mistake in the partial derivatives. Let me go back and check.
\end{tcolorbox}
\vspace{0.55em}
\begin{tcolorbox}[casegray]
\textbf{Wait}, so earlier, my mistake was in solving the equations. Upon correctly solving them, we find \(z=1\), \(x=\sqrt{2}\), \(y=\sqrt{2}\), which gives the minimal value of 16.
\end{tcolorbox}
\vspace{0.45em}
{\normalsize\bfseries Tokens: 20{,}974\quad \boxed{\boldsymbol{16}}}\ \cmark
\end{minipage}
\end{tcolorbox}
\caption{Gray is the shared prefix through the first same-answer window, and CALM continues in green after that window. The blue box opens with \texttt{</think>}. Both close at 16.}
\label{fig:case_fc_148}
\end{figure*}

\begin{figure*}[p]
\centering
\begin{tcolorbox}[caseouter, title={Question 165. Simplify \(\tan 100^\circ+4\sin 100^\circ\). Ground truth: \(-\sqrt{3}\).}]
\begin{minipage}[t]{0.48\linewidth}
\centering\textbf{\textsc{CALM}}\par\vspace{0.45em}
\begin{tcolorbox}[casegray]
So, \(\sin 100^\circ\,(1+4\cos 100^\circ)/\cos 100^\circ\).
\end{tcolorbox}
\vspace{0.55em}
\begin{tcolorbox}[casegray]
\textbf{Hmm}, perhaps I can write this as \(\tan 100^\circ\cdot(1+4\cos 100^\circ)\).
\end{tcolorbox}
\vspace{0.55em}
\begin{tcolorbox}[casegreen]
\(\tan 100^\circ\cdot(1+4\cos 100^\circ)\approx\tan 100^\circ\cdot(1+4\cdot(-0.173648178))\approx\tan 100^\circ\cdot(1-0.694592712)\approx\tan 100^\circ\cdot 0.305407288\).

But \(\tan 100^\circ\approx -5.67128182\), so
\[
-5.67128182\cdot 0.305407288\approx -1.73205081,
\]
which is approximately \(-\sqrt{3}\). So, that's consistent.
\end{tcolorbox}
\vspace{0.55em}
\begin{tcolorbox}[casegreen]
Let me compute \((\tan 100^\circ+4\sin 100^\circ)^2\).

\(\tan^2 100^\circ\approx 32.1653\), \(8\tan 100^\circ\sin 100^\circ\approx -44.672\), and \(16\sin^2 100^\circ\approx 15.517\).

Adding them together: \(32.1653-44.672+15.517\approx 3.000\).
\end{tcolorbox}
\vspace{0.55em}
\begin{tcolorbox}[casegreen]
So the square is about 3, and the sum itself is negative. Since \(100^\circ\) is in the second quadrant, tangent is negative and sine is positive. Therefore, it must be \(-\sqrt{3}\).
\end{tcolorbox}
\vspace{0.55em}
\begin{tcolorbox}[caseblue]
\texttt{</think>} The final answer is \(\boxed{-\sqrt{3}}\).
\end{tcolorbox}
\vspace{0.45em}
{\normalsize\bfseries Tokens: 4{,}702\quad \boxed{\boldsymbol{-\sqrt{3}}}}\ \cmark
\end{minipage}
\hfill
\begin{minipage}[t]{0.48\linewidth}
\centering\textbf{Full CoT}\par\vspace{0.45em}
\begin{tcolorbox}[casegray]
So, \(\sin 100^\circ\,(1+4\cos 100^\circ)/\cos 100^\circ\).
\end{tcolorbox}
\vspace{0.55em}
\begin{tcolorbox}[casegray]
\textbf{Hmm}, perhaps I can write this as \(\tan 100^\circ\cdot(1+4\cos 100^\circ)\).
\end{tcolorbox}
\vspace{0.55em}
\begin{tcolorbox}[casegray]
But we know that \(\tan 100^\circ\approx -5.671\), and \(1+4\cos 100^\circ\approx 0.3056\). Then \(\tan 100^\circ\cdot 0.3056\approx -1.732\), which is \(-\sqrt{3}\). So, the algebraic manipulation must lead to \(-\sqrt{3}\).
\end{tcolorbox}
\vspace{0.55em}
\begin{tcolorbox}[casegray]
\textbf{Alternatively}, maybe I can use some trigonometric identities to express this combination as a single sine or cosine function.
\end{tcolorbox}
\vspace{0.55em}
\begin{tcolorbox}[casegray]
\textbf{Wait}, another idea: Maybe use the identity for a linear combination of sine and tangent.
\end{tcolorbox}
\vspace{0.55em}
\begin{tcolorbox}[casegray]
But I'm not sure about that. \textbf{Alternatively}, perhaps use the sine of sum formula.
\end{tcolorbox}
\vspace{0.55em}
\begin{tcolorbox}[casegray]
\textbf{Wait}, let me consider the expression \(\tan\theta+4\sin\theta\). Let me write \(\tan\theta\) as \(\sin\theta/\cos\theta\), so it becomes \(\sin\theta\,(1+4\cos\theta)/\cos\theta\).
\end{tcolorbox}
\vspace{0.55em}
\begin{tcolorbox}[casegray]
\textbf{Alternatively}, perhaps I can consider that the expression is equal to \(-\sqrt{3}\), and then verify it using the approximate value, as I did before.
\end{tcolorbox}
\vspace{0.45em}
{\normalsize\bfseries Tokens: 13{,}134\quad \boxed{\boldsymbol{-\sqrt{3}}}}\ \cmark
\end{minipage}
\end{tcolorbox}
\caption{Gray is the shared prefix through the first same-answer window, and CALM continues in green after that window. The blue box opens with \texttt{</think>}. Both close at \(-\sqrt{3}\).}
\label{fig:case_fc_165}
\end{figure*}

\begin{figure*}[p]
\centering
\begin{tcolorbox}[caseouter, title={Question 146. \(z+1/z=(1+\sqrt{5})/2\). Find \(z^{85}+1/z^{85}\). Ground truth: \(-2\).}]
\begin{minipage}[t]{0.48\linewidth}
\centering\textbf{\textsc{CALM}}\par\vspace{0.45em}
\begin{tcolorbox}[casegray]
\textbf{Wait}, earlier, I computed \(a_3=-\phi+1\) directly, which is \(\frac{1-\sqrt{5}}{2}\). Let me compute \(\phi\cdot a_2-a_1\).
\end{tcolorbox}
\vspace{0.55em}
\begin{tcolorbox}[casegray]
\(a_2=\frac{\sqrt{5}-1}{2}\), so \(\phi\cdot a_2=\frac{1+\sqrt{5}}{2}\cdot\frac{\sqrt{5}-1}{2}\).

The numerators multiply to \(4\), so \(\phi\cdot a_2=\frac{4}{4}=1\).
\end{tcolorbox}
\vspace{0.55em}
\begin{tcolorbox}[casegreen]
Then, \(\phi\cdot a_2-a_1=1-\phi=\frac{1-\sqrt{5}}{2}\), which is indeed \(a_3\). So the correct recurrence is \(a_{n+1}=a_1 a_n-a_{n-1}\).
\end{tcolorbox}
\vspace{0.55em}
\begin{tcolorbox}[casegreen]
So, now, \(a_{10}=a_0\) and \(a_9=a_1\). That suggests that the sequence is periodic with period 10.
\end{tcolorbox}
\vspace{0.55em}
\begin{tcolorbox}[casegreen]
\(a_0=2\), \(a_1=\phi\), \(a_5=-2\), \(a_9=\phi\), \(a_{10}=2\).

Yes, from \(a_0\) to \(a_{10}\), the sequence repeats. Therefore, the period is 10.
\end{tcolorbox}
\vspace{0.55em}
\begin{tcolorbox}[casegreen]
Compute \(85\div 10=8\) with a remainder of 5. So \(85=10\times 8+5\). Therefore, \(a_{85}=a_5\).

From above, \(a_5=-2\).
\end{tcolorbox}
\vspace{0.55em}
\begin{tcolorbox}[caseblue]
\texttt{</think>} The final answer is \(\boxed{-2}\).
\end{tcolorbox}
\vspace{0.45em}
{\normalsize\bfseries Tokens: 5{,}904\quad \boxed{\boldsymbol{-2}}}\ \cmark
\end{minipage}
\hfill
\begin{minipage}[t]{0.48\linewidth}
\centering\textbf{DEER}\par\vspace{0.45em}
\begin{tcolorbox}[casegray]
Compute \(6\cdot 14=84\), so \(85-84=1\), and \(85\bmod 6=1\).

Therefore, \(a_{85}=a_1=\phi=(1+\sqrt{5})/2\).
\end{tcolorbox}
\vspace{0.55em}
\deertrial{\frac{1+\sqrt{5}}{2}}{\(<\lambda\)}
\vspace{0.55em}
\begin{tcolorbox}[casegray]
\textbf{Wait}, but earlier \(a_6=-2\), while \(a_0=2\). They are not the same. Is the period 12?

\textbf{Wait}, \(a_6=-2\), \(a_7=\phi\), \(a_{12}=2\). Thus \(85\bmod 12=1\), and \(a_{85}=a_1=\phi=(1+\sqrt{5})/2\).
\end{tcolorbox}
\vspace{0.55em}
\deertrial{\frac{1+\sqrt{5}}{2}}{\(<\lambda\)}
\vspace{0.55em}
\begin{tcolorbox}[casegray]
\textbf{Wait}, \(a_6=-2\) differs from \(a_0=2\), so the period is 12. Therefore \(a_n=a_{n\bmod 12}\), and the answer is \((1+\sqrt{5})/2\).
\end{tcolorbox}
\vspace{0.55em}
\deertrial{\frac{1+\sqrt{5}}{2}}{\(>\lambda\)}
\vspace{0.55em}
\begin{tcolorbox}[caseblue]
\texttt{</think>} \(12\cdot 7=84\), so \(85\bmod 12=1\), and \(a_{85}=a_1\).
\end{tcolorbox}
\vspace{0.55em}
\begin{tcolorbox}[caseblue]
The final answer is \(\boxed{\dfrac{1+\sqrt{5}}{2}}\).
\end{tcolorbox}
\vspace{0.45em}
{\normalsize\bfseries Tokens: 11{,}723\quad \boxed{\boldsymbol{\dfrac{1+\sqrt{5}}{2}}}}\ \xmark
\end{minipage}
\end{tcolorbox}
\caption{CALM is gray before the window and green after it, and DEER stays gray. The blue box opens with \texttt{</think>}. CALM closes at \(-2\), and DEER closes at \((1+\sqrt{5})/2\).}
\label{fig:case_deer_146}
\end{figure*}

\begin{figure*}[p]
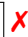

\centering
\begin{tcolorbox}[caseouter, title={Question 19. \(\overline{BC}\) is parallel to the segment through \(A\), and \(AB=BC\). How many degrees is \(x\)? Ground truth: 28.}]
\begin{minipage}[t]{0.48\linewidth}
\centering\textbf{\textsc{CALM}}\par\vspace{0.45em}
\begin{tcolorbox}[casegray]
The angle at \(A\) is given as 124 degrees, so if the angle at \(C\) were also 124 degrees, then \(124+124=248\), which is more than 180. So, that can't be.
\end{tcolorbox}
\vspace{0.55em}
\begin{tcolorbox}[casegray]
\textbf{Wait}, maybe it's triangle \(ABD\), where \(D\) is a point such that \(AD\) is parallel to \(BC\).

\textbf{Alternatively}, maybe the figure is a trapezoid with \(AB\) and \(CD\) as legs, and \(BC\) parallel to \(AD\), with \(AB=CD\).
\end{tcolorbox}
\vspace{0.55em}
\begin{tcolorbox}[casegreen]
Maybe I'm overcomplicating. Given: \(BC\) is parallel to a segment through \(A\), \(AB=BC\), and the angle at \(A\) is \(124^\circ\). The label for \(x\) is near \(AB\).
\end{tcolorbox}
\vspace{0.55em}
\begin{tcolorbox}[casegreen]
Since \(AB=BC\), triangle \(ABC\) is isosceles. The label for \(x\) is at \((4.5,3)\), on the top base, so perhaps \(x\) is the angle at \(B\) between \(AB\) and \(BC\).
\end{tcolorbox}
\vspace{0.55em}
\begin{tcolorbox}[casegreen]
In an isosceles trapezoid the base angles are equal. If the angle at \(A\) is \(124^\circ\), the angles at \(B\) and \(C\) are each \((180-124)/2=28^\circ\).
\end{tcolorbox}
\vspace{0.55em}
\begin{tcolorbox}[casegreen]
In the figure, \(x\) is labeled near \(AB\). So \(x\) is the angle at \(B\) between \(AB\) and \(BC\), which is \(28^\circ\).
\end{tcolorbox}
\vspace{0.55em}
\begin{tcolorbox}[caseblue]
\texttt{</think>} The final answer is \(\boxed{28}\).
\end{tcolorbox}
\vspace{0.45em}
{\normalsize\bfseries Tokens: 5{,}293\quad \boxed{\boldsymbol{28}}}\ \cmark
\end{minipage}
\hfill
\begin{minipage}[t]{0.48\linewidth}
\centering\textbf{DEER}\par\vspace{0.45em}
\begin{tcolorbox}[casegray]
Given angle \(BAD=124^\circ\), so angle \(ABC=56^\circ\).

In triangle \(ABC\), \(AB=BC\) and the angle at \(B\) is \(56^\circ\). The base angles are equal, so
\[
\text{angle }BAC+\text{angle }BAC+56=180.
\]
Thus angle \(BAC=62^\circ\). Therefore, angle \(BAC\) is 62 degrees. So, \(x\) is 62 degrees.
\end{tcolorbox}
\vspace{0.55em}
\deertrial{62}{\(<\lambda\)}
\vspace{0.55em}
\begin{tcolorbox}[casegray]
\textbf{Wait}, but let me verify this because I might have made a mistake.

\textbf{Wait}, angle \(BAC\) is part of angle \(BAD\). \textbf{Alternatively}, angle \(BCA\) is also 62 degrees. So angle \(x\) is 62 degrees.
\end{tcolorbox}
\vspace{0.55em}
\deertrial{62}{\(<\lambda\)}
\vspace{0.55em}
\begin{tcolorbox}[casegray]
\textbf{Wait}, but let me think again. Angle \(BAC\) is 62 and angle \(CAD\) is 62, so \(x\) is 62 degrees.
\end{tcolorbox}
\vspace{0.55em}
\deertrial{62}{\(>\lambda\)}
\vspace{0.55em}
\begin{tcolorbox}[caseblue]
\texttt{</think>} Sum of angles: \(56+2x=180\), so \(x=62\).
\end{tcolorbox}
\vspace{0.55em}
\begin{tcolorbox}[caseblue]
The final answer is \(\boxed{62}\).
\end{tcolorbox}
\vspace{0.45em}
{\normalsize\bfseries Tokens: 7{,}529\quad \boxed{\boldsymbol{62}}}\ \xmark
\end{minipage}
\end{tcolorbox}
\caption{CALM is gray before the window and green after it, and DEER stays gray. The blue box opens with \texttt{</think>}. CALM closes at 28, and DEER closes at 62.}
\label{fig:case_deer_19}
\end{figure*}

\section{AI Use Statement}
We utilized GPT-5.4 to assist in polishing the text, refining grammar, and improving the clarity of the manuscript. All content was reviewed and verified by the authors.